\documentclass[10.5pt]{article}
\usepackage{csquotes}

\usepackage[english]{babel}
\usepackage[a4paper,top=3.5cm,bottom=3.8cm,left=2.5cm,right=2.5cm,marginparwidth=1.75cm]{geometry}
\everymath{\displaystyle}

\newcommand{\Op}{\mathcal{G}^\dagger} 

\newcommand{\R}{\mathbb{R}} 
\newcommand{\Ca}{\text{Ca}^{2+}}

\newcommand{\Na}{\text{Na}}
\newcommand{\Cal}{\text{Ca}}
\newcommand{\K}{\text{K}}

\usepackage{amsmath}
\usepackage{amssymb}
\usepackage{mathtools}
\usepackage{microtype}
\usepackage{bbm}
\usepackage{graphicx}
\usepackage[colorlinks=true, allcolors=blue]{hyperref}
\usepackage{xcolor}
\usepackage{booktabs}
\usepackage{multicol}
\usepackage{subcaption}
\usepackage{enumitem}
\usepackage{tikz}
\usepackage[symbol]{footmisc}
\usetikzlibrary{positioning, arrows.meta, fit, shapes.geometric, calc}
\usepackage{todonotes}
\usepackage{authblk}
\usepackage{adjustbox}
\usepackage{tabularx}
\usepackage{subfiles} 
\newcommand{\learningRate}{\eta}

\newcommand{\schedulerGamma}{\gamma}
\newcommand{\weightDecay}{\omega}
\newcommand{\width}{d_v}
\newcommand{\hiddenLayer}{L}
\newcommand{\fourierModes}{k_{max}}
\newcommand{\activation}{\sigma}
\newcommand{\paddingPoints}{n_{pad}}

\title{Learning High-dimensional Ionic Model Dynamics\\ Using Fourier Neural Operators}

\author[1,2]{Luca Pellegrini\thanks{\textrm{luca.pellegrini02@universitadipavia.it}}}

\author[1]{Massimiliano Ghiotto\thanks{\textrm{massimiliano.ghiotto01@universitadipavia.it}}}
\author[1]{Edoardo Centofanti\thanks{\textrm{edoardo.centofanti01@universitadipavia.it}}}
\author[1]{Luca Franco Pavarino\thanks{\textrm{luca.pavarino@unipv.it}}}
\affil[1]{Department of Mathematics, University of Pavia, Pavia, Italy}
\affil[2]{Euler Institute, Faculty of Informatics, Università della Svizzera italiana, Lugano, Switzerland}
\date{}

\begin{document}




\maketitle

\hrulefill
\begin{abstract}
\noindent Ionic models, described by systems of stiff ordinary differential equations, are fundamental tools for simulating the complex dynamics of excitable cells in both Computational Neuroscience and Cardiology. Approximating these models using Artificial Neural Networks poses significant challenges due to their inherent stiffness, multiscale nonlinearities, and the wide range of dynamical behaviors they exhibit, including multiple equilibrium points, limit cycles, and intricate interactions. While in previous studies the dynamics of the transmembrane potential has been predicted in low dimensionality settings, in the present study we extend these results by investigating whether Fourier Neural Operators can effectively learn the evolution of all the state variables within these dynamical systems in higher dimensions. We demonstrate the effectiveness of this approach by accurately learning the dynamics of three well-established ionic models with increasing dimensionality: the two-variable FitzHugh-Nagumo model, the four-variable Hodgkin-Huxley model, and the forty-one-variable O'Hara-Rudy model. 
 
  To ensure the selection of near-optimal configurations for the Fourier Neural Operator, we conducted automatic hyperparameter tuning under two scenarios: an unconstrained setting, where the number of trainable parameters is not limited, and a constrained case with a fixed number of trainable parameters. Both constrained and unconstrained architectures achieve comparable results in terms of accuracy across all the models considered. However, the unconstrained architecture required approximately half the number of training epochs to achieve similar error levels, as evidenced by the loss function values recorded during training. These results underline the capabilities of Fourier Neural Operators to accurately capture complex multiscale dynamics, even in high-dimensional dynamical systems.
    \\[0.4cm]
\noindent\emph{Keywords}: \hspace{0.15cm}
    Fourier Neural Operator, Operator Learning, Numerical Machine Learning, Stiff Ionic Models, Computational Cardiology
\end{abstract}
\hrulefill



\section{Introduction}
Ionic models are an important class of dynamical systems describing the dynamics of excitable cells \cite{keener2009mathematical2}, finding applications across multiple fields such as Computational Neuroscience \cite{izhikevich2007dynamical} and Computational Cardiology \cite{franzone2014mathematical}. These complex dynamical systems, characterized by stiff ordinary differential equations (ODEs), exhibit rich dynamics, including multiple equilibrium points, limit cycles, and intricate interactions arising from nonlinearities across multiple timescales. Accurate solutions of these systems typically require multistep solvers tailored for stiff ODEs \cite{shampine1999solving}. The application of deep learning techniques to applied mathematics has gained significant attention in recent years, particularly using Artificial Neural Networks (ANNs) for solving both direct and inverse problems, which brought to the rapid development of Scientific Machine Learning \cite{cuomo2022scientific}. Among these techniques, Physics-Informed Neural Networks (PINNs) \cite{raissi2019physics} have gained interest by leveraging prior knowledge of the governing partial differential equations (PDEs) to optimize networks by directly imposing the underlying physics. While PINNs have demonstrated success in various fields, including Computational Neuroscience \cite{shekarpaz2024splitting}, they are typically trained to solve a single instance of a problem. Recently, an alternative paradigm based on ANNs, called Neural Operators (NOs) \cite{neuraloperator21kov}, has emerged, which aims to learn mappings between infinite-dimensional function spaces. These architectures are predominantly trained in a data-driven manner, although physics-based approaches can also be incorporated \cite{goswami2023physics}. Numerous NO architectures have been developed, including Deep Operator Networks (DeepONets) \cite{DON21lu}, Fourier Neural Operators (FNOs) \cite{FNO20li}, Wavelet Neural Operators (WNOs) \cite{tripura2023wavelet}, Convolutional Neural Operators  \cite{CNO23raonic}, Latent Dynamics Networks \cite{regazzoni2024learning}, and transformer-based architectures \cite{calvello2024continuum, li2022transformer, wu2024transolver}. 

A recent study \cite{centofanti2024learning} investigated the ability of DeepONets, FNOs, and WNOs to learn the mapping from applied current to voltage in the Hodgkin-Huxley model \cite{hodgkin1952quantitative}, and demonstrated that FNOs exhibit superior accuracy, particularly when the system presents a large number of oscillations. The present work further explores the capability of FNOs to learn the complex dynamics of multi-dimensional ionic models. In particular, the main novelties of this paper are:

\begin{itemize}

\item Learning all the variables of an ionic model simultaneously by a single FNO architecture, in contrast to previous approaches that either focused only on the transmembrane potential or employed separate architectures for each variable;

\item Extending this FNO architecture to high-dimensional ionic model, such as the O'Hara-Rudy model \cite{o2011simulation}, a forty-one variable model that provides a detailed description of human ventricular myocytes and is of particular interest in Computational Cardiology;

\item Exploring the dependence of the hyperparameters of the architecture employed on the dimensionality of the dynamical systems to learn;

\item  Applying state-of-the-art techniques of hyperparameters tuning on High Performance Computing (HPC) architectures and exploring the impact of parameter constraining in learning the solutions of the systems studied.

\end{itemize}

The paper is organized as follows: Section \ref{section:methodology} introduces Neural Operators, with a focus on Fourier Neural Operators, and defines the stiff ionic models under consideration: FitzHugh-Nagumo, Hodgkin-Huxley, and O'Hara-Rudy. Section \ref{section:numerical results} presents the numerical results. Then, in Section \ref{section:conclusions}, conclusions and future research directions are detailed.

\section{Mathematical models and methods}\label{section:methodology}
    Dynamical systems derived from ionic modeling play an important role in several fields \cite{franzone2014mathematical,izhikevich2007dynamical}. In particular, they can be described by system of ordinary differential equations of the following form:
    \begin{equation}\label{eq:stiff_ode}
            \begin{cases}
                    C_m\frac{dV}{dt} + I_{ion}(V,\vec{w},\vec{c}\,) =  I_{app},  &  t\in[0,T],                                            \\[12pt]
        
                    \frac{dw_j}{dt} = \alpha_j(V)(1-w_j) + \beta_j(V)w_j,\ \                                & j=1,...,M,    \\[12pt]
                    \frac{dc_k}{dt} = - \frac{I_{c_k}(V,w) A_{cap}}{V_{c_k} z_{c_k}}\,  F, \ \  & k =1,...,S,
                \end{cases}
    \end{equation}
    with initial conditions $V(0) = V_0,\ \ w_j(0) = w_{j,0}$ for all $j=1,...,M,\ \ c_k(0) = c_{k,0}$ for all $k=1,...,S$.
    Here the unknowns are the transmembrane potential $V$,  the vector of gating variables $\vec{w} = [w_1,\dots,w_M]$,  the vector of ionic concentration variables $\vec{c} = [c_1,\dots,c_S]$. We assume that the system coefficients, right-hand sides and initial conditions $V_0, \vec{w}_0, \vec{c}_0$ are such that \eqref{eq:stiff_ode} admits a unique solution $(V,\vec{w},\vec{c}\,)$.
    The aim of our work is to learn the operator that maps the applied current $I_{app}$ to the solution $(V,\vec{w},\vec{c}\,)$ of \eqref{eq:stiff_ode}:
    \begin{alignat}{2}\label{eq:operator_ion}
        \Op :  \R^+&\times [0,T]     &  & \rightarrow \mathcal{D}    \\
                   &I_{app} &  & \mapsto (V,\vec{w},\vec{c}\,) \notag
    \end{alignat}
    where the applied current is defined as a piecewise constant function
    \begin{center}
        $I_{app}(i,t) =
            \begin{cases}
                i, \ \ t\leq T_{stim}, \\
                0, \ \ t> T_{stim}.
            \end{cases}$
    \end{center}
   In this section, we introduce the necessary technical tools that will be used throughout this work. In particular, we begin with an introduction to Neural Operators, following the formalism presented in \cite{neuraloperator21kov}. We then define Fourier Neural Operators \cite{FNO20li}, that is a specific type of NO employed in this study. Afterwards, we give a more detailed description of ionic models \cite{franzone2014mathematical}. Finally, we discuss the three specific ionic models considered in this study: the FitzHugh-Nagumo \cite{fitzhugh1961impulses}, Hodgkin-Huxley \cite{hodgkin1952quantitative}, and O'Hara-Rudy \cite{o2011simulation} models.
  
    \subsection{Neural Operators}\label{section:neural_operators}
        Neural operators have emerged as a promising framework for learning mappings between infinite-dimensional function spaces, making them particularly well-suited for solving PDEs \cite{neuraloperator21kov}. Unlike traditional neural networks that operate on finite-dimensional spaces, neural operators can learn continuous operators that map between function spaces while maintaining independence from the discretization resolution. This section outlines the key components of Neural Operators, based on the work of Kovachki et al. \cite{Kamyar, neuraloperator21kov}. Consider a class of partial differential equations defined on a bounded spatio-temporal domain \( D \subset \mathbb{R}^d \) with boundary \( \partial D \). The problem can be formulated as:
        \begin{equation*}
            \begin{cases}
                (\mathcal{N}_a u)(x) = f(x), \quad &x \in D,          \\
                (\mathcal{B}u)(x)    = g(x), \quad &x \in \partial D.
            \end{cases}
        \end{equation*}
        Here, $\mathcal{N}_a : \mathcal{U}(D, \R^{d_u}) \rightarrow \mathcal{U}(D, \R^{d_u})^* $ is a differential operator parameterized by $a\in \mathcal{A}(D, \R^{d_a})$ and $\mathcal{B}$ is the boundary operator. In this setting, $\mathcal{U}(D, \R^{d_u})$ and $\mathcal{A}(D, \R^{d_a})$ are Sobolev spaces and $\mathcal{U}(D, \R^{d_u})^*$ is the dual space of $\mathcal{U}(D, \R^{d_u})$. We define the solution operator $\mathcal{G}^\dagger := \mathcal{N}_a^{-1}$, which maps the parameters $a$ to the solution $u$, establishing the correspondence $\mathcal{G}^\dagger: \mathcal{A}(D, \R^{d_a}) \to \mathcal{U}(D, \R^{d_u})$. The fundamental objective of NOs is to develop a data-driven approximation for the solution operator $\Op$. In general, NOs are defined as a composition of operators:
        \begin{equation*}
            \mathcal{G}_{\theta} :\mathcal{A}(D, \R^{d_a}) \to \mathcal{U}(D, \R^{d_{u}}), \quad	\mathcal{G}_{\theta} := \mathcal{Q} \circ \mathcal{L}_L \circ \cdots \circ \mathcal{L}_1 \circ \mathcal{P},
        \end{equation*}
        where the three main components are:
        \begin{itemize}
            \item a lifting operator $\mathcal{P}:\, \mathcal{A}(D, \R^{d_a}) \to \mathcal{U}(D, \R^{d_{v}}),\  v_0(x) \coloneqq \mathcal{P}(a)(x)\in \R^{d_v},$ with the choice $d_v > d_a$, which maps the input function to a higher-dimensional space. This operator is typically implemented either as a shallow pointwise Multilayer Perceptron (MLP) or a single linear layer;
            \item a composition of integral operators $ \mathcal{L}_t : \, \mathcal{U}(D, \R^{d_{v}}) \to  \mathcal{U}(D, \R^{d_{v}}),\ v_{t}(x) = \mathcal{L}_t(v_{t-1})(x)\in \R^{d_v}, $\\ $t=1,\ldots,L$, processing the lifted representation through \( L \) layers in order to capture global dependencies and interactions in the input function through a sequence of transformations. Each $\mathcal{L}_t$ has the following structure:
            \begin{equation}\label{eq:integral_operator_fno}
                \mathcal{L}_t(v)(x) := \sigma\Big( W_t v(x)+ b_t + \mathcal{K}_t(a, \theta_t) (v(x)) \Big),
            \end{equation}
            where $W_{t} \in \R^{d_{v} \times d_v}$,$\ b_{t} \in \R^{d_v}$, $\theta_t \in \Theta_t \subset \Theta$ is a subset of the trainable parameters, and $\sigma$ is a non-linear activation function;
            \item a projection operator $\mathcal{Q}:\, \mathcal{U}(D, \R^{d_{v}}) \to  \mathcal{U}(D, \R^{d_{u}}),\ u(x) = \mathcal{Q}(v_L)(x)\in \R^{d_u}, $ which maps the internal representation back to the output space. This operator is also implemented as a pointwise MLP.
        \end{itemize}
        The ability of neural operators to effectively approximate PDE solution operators is related to some relevant theoretical properties. In particular, the composition of the integral layers must exhibit both non-linearity and non-locality to accurately capture the complex dynamics of the underlying PDEs \cite{lanthaler2023nonlocal}. Furthermore, the architecture must maintain mesh resolution independence, ensuring that the learned operator generalizes across different discretizations of the domain \cite{bartolucci2023neural}.
        
        \subsubsection{Fourier Neural Operator}\label{subsection:fno}
            Fourier Neural Operators \cite{neuraloperator21kov, FNO20li} are a class of neural operators that employs the Fourier transform to efficiently parameterize the integral operator $\mathcal{L}_t$ \eqref{eq:integral_operator_fno}. By operating in the frequency domain, FNOs effectively capture global dependencies within input functions. In the case of FNO, the integral kernel operator $\mathcal{K}_t(a, \theta_t)$ included in \eqref{eq:integral_operator_fno} has the form:
            \begin{equation}\label{eq:kernel}
                (\mathcal{K}_t(a, \theta_t)v)(x) = (\mathcal{K}_t(\theta_t)v)(x) = \int_{\mathbb{T}^d} \kappa_{t,\theta_t}(x,y) v(y) \ dy = \int_{\mathbb{T}^d} \kappa_{t,\theta_t}(x-y) v(y) \ dy= (\kappa_{t, \theta_t} * v)(x) ,
            \end{equation}    
            where $\mathbb{T}^d$ is the $d$-dimensional torus, and $\kappa_{t,\theta_t}$ is a kernel function, which depends on the learnable parameters $\theta_t$. 
            We can apply the convolution theorem for the Fourier transform in order to rewrite the convolution in \eqref{eq:kernel} in the Fourier domain:
            \begin{equation*}
             (\kappa_{t, \theta_t} * v)(x) =  \mathcal{F}^{-1}\left( \mathcal{F}( \kappa_{t,\theta_t}) (k) \cdot \mathcal{F}(v)(k) \right)(x).
            \end{equation*}
            The key step in FNO is the parameterization of $\mathcal{F}( \kappa_{t, \theta_t} )(k)$ using a complex-valued matrix of learnable parameters $R_{\theta_t}(k) \in \mathbb{C}^{d_v \times d_v}$ for each frequency mode $k \in \mathbb{Z}^d$, obtaining:
            \begin{equation*}
                 (\mathcal{K}_t(\theta_t)v)(x)= \mathcal{F}^{-1}\left( R_{\theta_t}(k) \cdot \mathcal{F}(v)(k) \right)(x). 
            \end{equation*}
            Since we consider real-valued functions, the parameterized matrix $R_{\theta_t}(k)$ must be Hermitian, i.e., $R_{\theta_t}(-k) = \overline{R_{\theta_t}(k)}$ for all $k \in \mathbb{Z}^d$ and $t=1,\ldots,L$. Furthermore, in order to apply the Fast Fourier Transform (FFT) for the numerical implementation of the Fourier transform, the mesh considered must be structured and uniform.
            \begin{figure}[!ht]
                \centering
                \begin{tikzpicture}
                    \tikzset{
                        box/.style={draw, rounded corners, align=center, minimum height=0.8cm, minimum width=1cm, fill=orange!31},
                        bigbox/.style={draw, rounded corners, align=center, minimum height=2cm, minimum width=1cm, fill=yellow!31},
                        node_sum/.style={draw, circle, fill=white, inner sep=0pt, minimum size=4mm},
                        every node/.style={font=\small}
                    }
                    \node[box, label=above:{\textit{Input}}] (input) {$a(x)$};
                    \node[box, right=0.5cm of input, label=above:{\textit{Lifting}}] (lifting) { $\mathcal{P}$ };
                    \node[bigbox, right=0.5cm of lifting] (fourier1) {$\mathcal{L}_{1}$};
                    \node[bigbox, right=0.55cm of fourier1] (fourier2) {$\mathcal{L}_{t}$};
                    \node[bigbox, right=0.55cm of fourier2] (fourier3) {$\mathcal{L}_{L}$};
                    \node[box, right=0.5cm of fourier3, label=above:{\textit{Projection}}] (projection) {$\mathcal{Q}$};
                    \node[box, right=0.5cm of projection, label=above:{\textit{Output}}] (output) {$u(x)$};
            
                    \draw[-stealth, line width = .7pt] ($(input.east)+(0.05, 0)$) -- ($(lifting.west)-(0.05,0)$);
                    \draw[-stealth, line width = .7pt] ($(lifting.east)+(0.05, 0)$) -- ($(fourier1.west)-(0.03,0)$);
                    \draw[dotted, line width = 2pt] ($(fourier1.east)+(0.1, 0)$) -- ($(fourier2.west)-(0.1,0)$);
                    \draw[dotted, line width = 2pt] ($(fourier2.east)+(0.1, 0)$) -- ($(fourier3.west)-(0.1,0)$);
                    \draw[-stealth, line width = .7pt] ($(fourier3.east)+(0.05, 0)$) -- ($(projection.west)-(0.05,0)$);
                    \draw[-stealth, line width = .7pt] ($(projection.east)+(0.05, 0)$) -- ($(output.west)-(0.05,0)$);
            
                    \node[align=center, above=0.2cm of fourier2, font=\footnotesize] { \textit{Fourier Layers} };
            
                    \node[draw, line width = .7pt, rounded corners, inner sep=0.2cm, fit= (fourier1) (fourier2) (fourier3)] (internal) {};
            
                    \node[draw, below=0.5cm of internal, rounded corners, inner sep=0.2cm, fill=yellow!30] (internal) {
                        \begin{tikzpicture}[every node/.style={font=\small}]
                            \node[box] (vt) {$v_t(x)$};
                            \node[box, right=1cm of vt] (transform) {$\mathcal{F}$};
                            \node[box, right=0.5cm of transform, fill=green!26] (nonlinear) { $ R_{\theta_t} $ };
                            \node[box, right=0.5cm of nonlinear] (invtransform) { $ \mathcal{F}^{-1} $ };
                            \node[box, below=0.8cm of nonlinear, fill=green!26] (linear) { $ W_t, b_t $ };
            
                            \node[draw, line width = .7pt, rounded corners, inner sep=0.15cm, fit= (transform) (nonlinear) (invtransform) ] (diagonalscaling) {};
            
                            \node[node_sum, right=0.5cm of invtransform] (node_sum) {$\mathbf{+}$};
                            \node[box, right=0.5cm of node_sum] (activation) {$\sigma$};
                            \node[box, right=0.5cm of activation] (vtplusone) {$v_{t+1}(x)$};
            
                            \draw[line width = .7pt] ($(vt.east)+(0.05, 0)$) -- (diagonalscaling);
                            \draw[-stealth, line width = .7pt] ($(transform.east)+(0.05, 0)$) -- ($(nonlinear.west)-(0.05, 0)$);
                            \draw[-stealth, line width = .7pt] ($(nonlinear.east)+(0.05, 0)$) -- ($(invtransform.west)-(0.05, 0)$);
                            \draw[-stealth, line width = .7pt] ($(vt.south)-(0, 0.05)$) |- ($(linear.west)-(0.05, 0)$);
                            \draw[-stealth, line width = .7pt] (diagonalscaling) -- ($(node_sum.west)-(0.05, 0)$);
                            \draw[-stealth, line width = .7pt] ($(linear.east)+(0.05, 0)$) -| ($(node_sum.south)-(0, 0.05)$);
                            \draw[-stealth, line width = .7pt] ($(node_sum.east)+(0.05, 0)$) -- ($(activation.west)-(0.05, 0)$);
                            \draw[-stealth, line width = .7pt] ($(activation.east)+(0.05, 0)$) -- ($(vtplusone.west)-(0.05, 0)$);
            
                        \end{tikzpicture}
                    };
                    \draw[] ($(internal.north west)+(0.1, 0.05)$) -- (fourier2.south west);
                    \draw[] ($(internal.north east)+(-0.1, 0.05)$) -- (fourier2.south east);
                \end{tikzpicture}
                \label{fig:fno}
                \caption{Visual representation of a Fourier Neural Operator.}
            \end{figure}
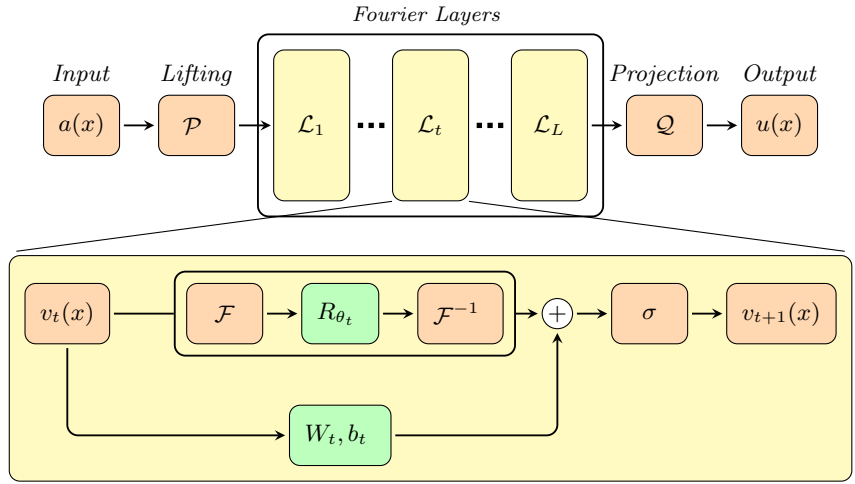
        
        In our implementation, we have considered two versions of \eqref{eq:integral_operator_fno}:
        \begin{align}
                v_{t+1}^{Classic} &= \mathcal{L}_t^{Classic}(v_t) := \sigma\Big( W_t v_t+ b_t + \mathcal{K}_t(\theta_t) v_t \Big), \label{eq:classic} \\
                v_{t+1}^{MLP} &= \mathcal{L}_t^{MLP}(v_t) := \sigma\Big( W_t v_t+ b_t + MLP(\mathcal{K}_t(\theta_t) v_t )\Big) \label{eq:mlp},
        \end{align} 
       where $v_{t+1}^{\text{Classic}}$ has the exact structure described in \eqref{eq:integral_operator_fno}, while
       $v_{t+1}^{\text{MLP}}$ generalizes the classic architecture by incorporating a pointwise MLP applied to the output of the integral kernel operator.

    \subsection{Stiff ionic models}\label{section:Stiff Ionic Models}
    Cell excitability is a fundamental property of various cells, including neurons \cite{izhikevich2007dynamical}, muscle cells \cite{keener2009mathematical2} and cardiomyocytes \cite{franzone2014mathematical}.
   A key characteristic of these excitable cells is the threshold effect: 
   if the stimulating current is below a critical combination of intensity and duration, the transmembrane potential quickly returns to its resting value after the stimulus ends; however, if the stimulating current exceeds this critical combination, the transmembrane potential undergoes a very rapid and large change (known as action potential) before returning to its resting value after the stimulus ends. 
   The time evolution of the transmembrane potential in these systems is typically  modeled by a system of stiff ordinary differential equations \eqref{eq:stiff_ode}. 
The total ionic current $I_{ion}$ flowing across the membrane is defined as:
            \begin{center}
                $I_{ion}(V,\vec{w},\vec{c}) = \sum_{k=1}^{N} G_k(V,\vec{c}\,)\prod_{j=1}^{M} w_k^{p_{j_k}}(V - V_k(\vec{c}\,)) + I_n(V,\vec{w},\vec{c}\,)$.
            \end{center}    
    Here, $N$ is the number of ionic currents, $G_k$ is the membrane conductance, $V_k$ is called the \textit{reversal potential} for the k-th ionic current, $p_{j_k}$  are integers, and $I_n$ accounts for time independent ionic fluxes. The number of equations within this system varies depending on the specific ionic model being considered \cite{franzone2014mathematical, izhikevich2007dynamical, keener2009mathematical2}. 
    To evaluate the capability of FNOs to accurately capture these complex dynamics, we examined three well-established ionic models with increasing dimensionality: the FitzHugh-Nagumo \cite{fitzhugh1961impulses}, Hodgkin-Huxley \cite{hodgkin1952quantitative}, and O'Hara-Rudy \cite{o2011simulation} models, which are briefly described in the following sections.
    
    \subsubsection{FitzHugh-Nagumo model}\label{subsection:fhn}
    The FitzHugh-Nagumo model \cite{fitzhugh1961impulses} is a two-variable dynamical system. It can be derived from an electric circuit or from a simplification of the Hodgkin-Huxley model. It is represented by the following system:
        \begin{equation}\label{equation:fhn}
        \begin{cases}
                \frac{dV}{dt} = bV(V-\beta)(\delta-V) -cw +I_{app}, \ \  & t\in[0,T],  \\[12pt]
    
                \frac{dw}{dt} = e(V-\gamma w).             
            \end{cases}
        \end{equation}    
    Here, $V$ is the transmembrane potential and $w$ acts as a recovery variable representing the activation of sodium channels and deactivation of potassium channels after stimulation by an external input current. This model can be viewed as a reduced and computationally efficient two-dimensional form of the more complex four-dimensional Hodgkin-Huxley model. It provides a valuable framework for exploring a wide range of current-induced dynamic behaviors. Although it does not explicitly model detailed subcellular mechanisms, its simplicity has allowed for extensive phase-plane analyses in the literature, see e.g. \cite{cebrian2024six,franzone2014mathematical}.
    
    \subsubsection{Hodgkin-Huxley model}\label{subsection:hh}
     The Hodgkin-Huxley model, developed by Alan Hodgkin and Andrew Huxley in 1952 \cite{hodgkin1952quantitative}, was originally introduced to accurately describe the action potential in the giant axon of the squid. It has since become the prototype for mathematical modeling of excitable cells. The model is defined by a system of four ODEs:
        \begin{equation}\label{equation:hh}
            \begin{cases}
                C_m\frac{dV}{dt} + I_{ion}(V,m,h,n) = I_{app}, \ \  & t\in[0,T], \\[12pt]
            
                \frac{dm}{dt} = \alpha_m(V)(1-m)-\beta_m(V)m, \\[12pt]
                \frac{dh}{dt} = \alpha_h(V)(1-h)-\beta_h(V)h, \\[12pt]
                \frac{dn}{dt} = \alpha_n(V)(1-n)-\beta_n(V)n.
            \end{cases}
        \end{equation}
    Here $V$ is the transmembrane potential, $m$ and $h$ are the gating variables that govern the activation and inactivation of sodium channels, respectively, $n$ is the gating variable for potassium channels, and $I_{ion}$ is the sum of three primary ionic currents: the sodium current ($I_{\Na}$), the potassium current ($I_{\K}$), and the leak current ($I_L$), as described by:
    \begin{equation*}
        I_{ion} =  \underbrace{\bar{g}_{\Na}m^3h(V-V_{\Na})}_{I_{\Na}} + \underbrace{\bar{g}_\K n^4(V-V_\K )}_{I_\K }+\underbrace{\bar{g}_L(V-V_L)}_{I_L}.
    \end{equation*}
    The Hodgkin-Huxley model paved the way for a variety of subsequent models, ranging from extensions incorporating additional ion channels and currents, as used in Computational Cardiology (e.g., the O'Hara-Rudy model \cite{o2011simulation}), to simplifications designed for efficient simulation of large neural networks in Computational Neuroscience (e.g., the FitzHugh-Nagumo model \cite{fitzhugh1961impulses}).
    
    \subsubsection{O'Hara-Rudy model}\label{subsection:Ord}
    The intricate dynamics of intracellular calcium plays a fundamental role in cardiac electrophysiology, not only governing the excitation-contraction coupling that drives the heart's pumping action, but also influencing the electrical signals that coordinate its rhythm. Impaired calcium handling is implicated in a wide range of cardiac pathologies, including arrhythmias, heart failure, and ischemia, making its accurate representation in computational models essential. Models based on non-human myocytes can yield results that are affected by differences between cell types and species. Therefore, an accurate model based on human cells is critical for understanding cardiac arrhythmias and related human pathologies. In 2011, O'Hara and Rudy developed a detailed ionic model of human ventricular myocytes \cite{o2011simulation}. This system of ordinary differential equations consists of forty-one variables, with a total ionic current $I_{ion}$ given by the sum of fourteen individual currents:
    \begin{equation}\label{eq:ord_curr}
    I_{ion} = I_{\Na} + I_{to} + I_{\Cal L} + I_{\Cal \Na} + I_{\Cal \K} + I_{\K r} + I_{\K s} + I_{\K 1} + I_{\Na \Cal } + I_{\Na \K} + I_{\Na b} + I_{\Cal b} + I_{\K b} + I_{p\Cal }.
    \end{equation}
    We refer to Table~\ref{table:Iion-notation} for a description of each term in \eqref{eq:ord_curr} and to the original paper \cite{o2011simulation} for all the model equations.
    \begin{table}[h!]
    \centering
    \begin{tabular}{ll}
        \hline
        \textbf{Current} & \textbf{Descrip}\textbf{tion} \\
        \hline\hline
        $I_{\Na}$ & Na$^+$ current \\
        $I_{to}$ & Transient outward K$^+$ current \\
        $I_{\Cal L}$ & Ca$^{2+}$ current through the L-type Ca$^{2+}$ channel \\
        $I_{\Cal \Na}$ & Na$^+$ current through the L-type Ca$^{2+}$ channel \\
        $I_{\Cal \K}$ & K$^+$ current through the L-type Ca$^{2+}$ channel \\
        $I_{\K r}$ & Rapid delayed rectifier K$^+$ current \\
        $I_{\K s}$ & Slow delayed rectifier K$^+$ current \\
        $I_{\K 1}$ & Inward rectifier K$^+$ current \\
        $I_{\Na \Cal}$ & Total Na$^+$/Ca$^{2+}$ exchange current \\
        $I_{\Na \K}$ & Na$^+$/K$^+$ ATPase current \\
        $I_{\Na b}$ & Na$^+$ background current \\
        $I_{\Cal b}$ & Ca$^{2+}$ background current \\
        $I_{\K b}$ & K$^+$ background current \\
        $I_{p\Cal}$ & Sarcolemmal Ca$^{2+}$ pump current \\
        \hline
    \end{tabular}
    \caption{Description of the individual ionic currents contributing to $I_{ion}$ in the O'Hara-Rudy model \cite{o2011simulation}.}
    \label{table:Iion-notation}
\end{table}

\section{Numerical results}\label{section:numerical results}
    In this section, we present the results of the numerical test obtained using Fourier Neural Operators to approximate \eqref{eq:operator_ion} for each of the test cases presented in Section \ref{section:Stiff Ionic Models}. For our numerical tests, we employ HyperNOs \cite{ghiotto2025hypernosautomatedparallellibrary} a \texttt{PyTorch} based library that employs \texttt{Ray} \cite{liaw2018tune} for the Hyperparameter tuning. All tests are performed on a workstation equipped with a single NVIDIA RTX 4090 GPU, while the hyperparameters search for the O'Hara-Rudy case has been performed on the CINECA cluster LEONARDO \cite{cineca_leonardo}, employing up to 4 
    nodes, each one equipped with one Intel Xeon Platinum 8358 (32 cores) CPU and 4 NVIDIA A100 64GB GPUs.

\subsection{Automated hyperparameter tuning}
    Optimization of hyperparameters is crucial for achieving high accuracy when employing Artificial Neural Network architectures. In our study, we considered the hyperparameters tuning for the Fourier Neural Operator for all the parameters listed in Table \ref{table:hyperparams}.
    \begin{table}[h!]
        \centering
        \begin{tabular}{ll}
            \hline
            \textbf{Hyperparameter} & \textbf{Description} \\
            \hline\hline
            $\width$ & Hidden dimension \\
            $\hiddenLayer$ & Number of hidden layers \\
            $\fourierModes$ & Number of Fourier modes \\
            $\activation$ & Activation function \\
            $\paddingPoints$ & Padding points \\
            Classic or MLP & Fourier architecture \eqref{eq:classic}, \eqref{eq:mlp}  \\
            $\weightDecay$ & Weight decay regularization factor \\
            $\learningRate$ & Learning rate \\
            $\schedulerGamma$ & Rate scheduler factor \\
            \hline
        \end{tabular}
        \caption{FNO hyperparameters that were optimized.}
        \label{table:hyperparams}
    \end{table}
    
    We performed 200 trials employing automatic hyperparameter optimization using HyperOptSearch, through the Tree-structured Parzen Estimator algorithm \cite{hyperopt11bergstra}, and ASHA scheduler \cite{AHSHA} for automatic stopping for not satisfactory trials. Each model was trained for 1000 epochs with mini-batches of 32 samples, using  as loss function the relative $L^2$ norm  defined as follows:
\[
    \begin{split}
        \text{Loss}\left( [u_j]_{j = 1}^{n_{dim}},\, \mathcal{G}_\theta(a)\right) & = \frac{1}{n_{dim}} \sum_{j=1}^{n_{dim}}\frac{\|u_j - \mathcal{G}_\theta(a)_j\|_{L^2(D)}}{\|u_j\|_{L^2(D)}}                                                                                                       \\
         & \approx \frac{1}{n_{dim}} \sum_{j=1}^{n_{dim}} \frac{\left(\sum_{k=1}^{n_{points}} |u_j(x_k) - \mathcal{G}_\theta(a)_j(x_k)|^2\right)^{1/2}}{\left(\sum_{k=1}^{n_{points}} |u_j(x_k)|^2\right)^{1/2}},
    \end{split}
\]
where $n_{dim}$ is the number of variables of the system. This leads to the following optimization problem:
\[
    \underset{\theta\in\Theta}{\arg\min} \ \ \frac{1}{n_{train}} \sum_{i=1}^{n_{train}} \text{Loss}\left( [u^i_j]_{j = 1}^{n_{dim}},\, \mathcal{G}_\theta(a^i)\right).
\]

     The optimizer we consider for this minimization problem is \texttt{AdamW} \cite{loshchilov2017decoupled} with weight decay regularization $\weightDecay$. We also employ a learning rate scheduler that reduces the learning rate $\learningRate$ by a factor of $\schedulerGamma$ every $10$ epochs. The dataset was partitioned into training, validation, and test sets with an 80/10/10 split:
\begin{itemize}
    \item Training set ($80\%$ of the total dataset): used to train our model.
    \item Validation set ($10\%$ of the total dataset):  used for hyperparameter selection.
    \item Test set (the remaining $10\%$ of the total dataset): the model with the lower validation error is finally tested on this part of the dataset and the test errors are reported in this paper in Tables~\ref{table:numerical_results_fhn_fno}, \ref{table:numerical_results_hh_fno}, and \ref{table:numerical_results_ord_fno}.
\end{itemize}
Furthermore, two hyperparameter search strategies were employed: \textit{unconstrained optimization} and \textit{constrained optimization}. The unconstrained optimization does not impose any limit on the number of trainable parameters, while the constrained optimization limits to approximately 500,000 the number of trainable parameters. The optimal hyperparameters resulting from the unconstrained search are presented in Table \ref{table:hyper-FNO-uncostrained}, and the results from the constrained optimization are shown in Table \ref{table:hyper-FNO-samedofs}. 
\begin{table}[ht!]
    \centering
    \begin{tabular}{cccccccccc}
        \hline
        Model & $\learningRate$ & $\weightDecay$ & $\schedulerGamma$ & $\width$ & $\hiddenLayer$ & $\fourierModes$ & $\activation$ & $\paddingPoints$ & FNO\_mod \\
        \hline\hline
        FHN     & 6.2e-4     & 9.7e-4    & 0.85            & 224    & 4            & 20             & leaky relu   & 15              & MLP \\
        HH      & 4.4e-4     & 7.5e-4      & 0.88            & 256    & 4            & 24             & leaky relu   & 15              & MLP   \\
        ORd      & 8.3e-4     & 1e-3      & 0.93            & 192    & 5            & 32             & gelu  & 14              & MLP   \\
        \hline
    \end{tabular}
    \caption{Hyperparameters configuration from the unconstrained optimization process over the configuration space.}
    \label{table:hyper-FNO-uncostrained}
\end{table}
\begin{table}[ht!]
    \centering
    \begin{tabular}{cccccccccc}
        \hline
        Model & $\learningRate$ & $\weightDecay$ & $\schedulerGamma$ & $\width$ & $\hiddenLayer$ & $\fourierModes$ & $\activation$ & $\paddingPoints$ & FNO\_mod \\
        \hline\hline
        FHN     & 1.7e-3     & 4.8e-4    & 0.93            & 64    & 5            & 12             & gelu   & $5$              & MLP \\
        HH      & 7.6e-3     & 5.1e-4      & 0.92            & 96    & 5            & 5             & gelu   & 7              & MLP   \\
        ORd      & 4.2e-3     & 4.9e-4     & 0.93            & 32    & 5            & 48             & gelu   & 15              & Classic   \\
        \hline
    \end{tabular}
    \caption{Hyperparameter configuration from the constrained optimization process over the configuration space consisting of approximately $500,000$ trainable parameters.}
    \label{table:hyper-FNO-samedofs}
\end{table}

\subsection{FitzHugh-Nagumo results}
We first considered the FitzHugh-Nagumo model, which was introduced in Section \ref{subsection:fhn}. The parameters considered in Equation \eqref{equation:fhn} are given in Table \ref{table:fhn parameters}.
\begin{table}[ht!]
    \centering
    \begin{tabular}{c|ccccccccc}
        \hline
        Parameter & b  & $\beta$ & c & $\delta$ & $\gamma$ & e & T (ms)& $V_0$ & $w_0$ \\
        \hline 
        Value & 5  & 0.1 & 1 & 1 & 0.25 & 1 & 100 & 0 & 0 \\
        \hline
    \end{tabular}
    \caption{FHN parameters, see \eqref{equation:fhn}.}
    \label{table:fhn parameters}
\end{table}

The dataset, generated using a Runge-Kutta method for stiff ODEs \cite{shampine1999solving}, consists of 3000 training, 375 test, and 375 for validation examples. To ensure a comprehensive coverage of the system dynamics, the training dataset was divided into three subsets. Each subset was generated by uniformly sampling current intensity ($i$) and stimulus duration ($T_{stim}$) within specific ranges. The training, validation and test datasets are described in detail in Table \ref{table:fhn_dataset}.

\begin{table}[ht!]
    \centering
    \begin{tabular}{cccccc}
        \hline
        \multicolumn{6}{c}{\textbf{Training Dataset}} \\
        \hline
        Name    & \multicolumn{2}{c}{Range of values of $i$} & \multicolumn{2}{c}{Range of values of $T_{stim}$} & Number of examples              \\
                & min                                        & max                                               & min                & max &      \\

        \hline\hline
        $t_0$   & -                                          & -                                                 & 0                  & 0   & 20   \\
        General & 0.1                                        & 2                                                 & 0.01               & 100 & 2480 \\
        nap     & 1e-4                                       & 0.01                                              & 0.01               & 100 & 500  \\
        \hline
        \multicolumn{6}{c}{} \\
        \hline
        \multicolumn{6}{c}{\textbf{Test and Validation Dataset}} \\
        \hline
        Name      & \multicolumn{2}{c}{Range of values of $i$} & \multicolumn{2}{c}{Range of values of $T_{stim}$} & Number of examples             \\
                  & min                                        & max                                               & min                & max &     \\

        \hline\hline
        $t_0$     & -                                          & -                                                 & 0                  & 0   & 10  \\
        General   & 0.1                                        & 2                                                 & 0.01               & 100 & 285 \\
        nap       & 1e-4                                       & 0.01                                              & 0.01               & 100 & 30  \\
        $t_{fin}$ & 0.3                                        & 3                                                 & 100                & 100 & 50  \\
        \hline
    \end{tabular}
    \caption{FHN model: Range of values of the training set (first table) and the test and validation dataset (second table). The training dataset is divided into three subsets, while the test and validation datasets are divided into four subsets. Specifically, we introduced a new subset called $t_{fin}$, where the stimulus is applied until the final simulation time.}
    \label{table:fhn_dataset}
\end{table}

The hyperparameters explored for the unconstrained and constrained cases are presented in Table \ref{table:hyper-FNO-uncostrained} and Table \ref{table:hyper-FNO-samedofs}, respectively. Table \ref{table:numerical_results_fhn_fno} summarizes the performance of these architectures for the FHN case, including the number of trainable parameters (in millions), memory usage during training (GB), training time (minutes), and the mean relative $L^2$ error. The latter is obtained by averaging over five runs, each one with different random seeds for the initialization of the architecture weights.
\begin{table}[ht!]
    \centering
    \begin{tabular}{ccccc|c}
        \hline
        Mode       & Number of & Memory & Time & \multicolumn{2}{c}{Rel. $L^2$ error}                         \\
                   &       parameters   &              &               & Training                            & Test                  \\
        \hline\hline
        constr. optim. & $0.57$ M   & $0.98$ GB    & $19$ min     & $(0.35 \pm 0.009)\%$                & $(0.86 \pm 0.042) \%$ \\
        unconstr. optim.      & $8.69$ M   & $1.76$ GB   & $31$ min     & $(0.31 \pm 0.004) \%$               & $(0.87 \pm 0.014) \%$ \\
        \hline
    \end{tabular}
    \caption{FHN model: FNO result summary.}
    \label{table:numerical_results_fhn_fno}
\end{table}

Regarding the results reported in Table \ref{table:numerical_results_fhn_fno}, we can observe that the unconstrained architecture did not offer a significant accuracy advantage over the constrained architecture. To further evaluate and investigate the differences between these two architectures we examined the relative $L^1$, $L^2$, and $H^1$ norms, averaged over five runs with different random seeds for weight initialization.
\begin{figure}[ht]
    \centering
    \begin{subfigure}{0.48\textwidth}
        \centering
        \includegraphics[width=\textwidth]{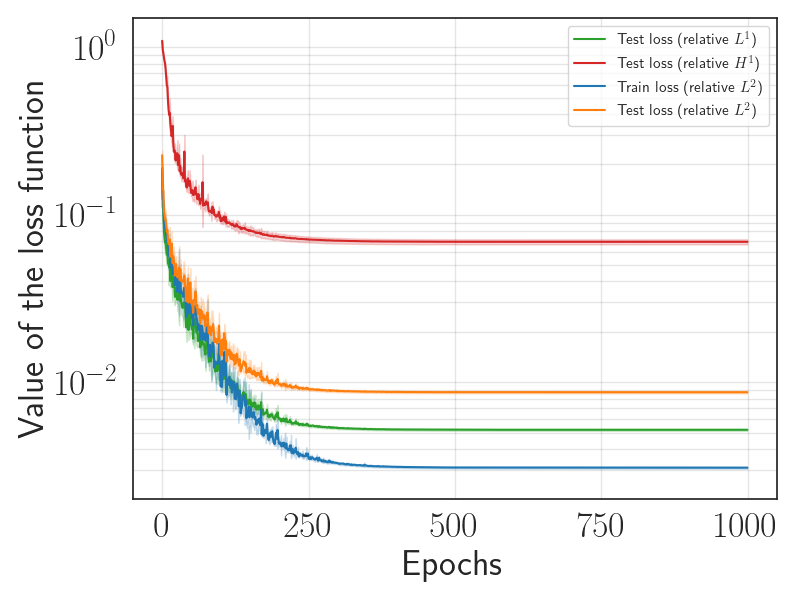}
        \caption{Loss values for the unconstrained FNO.}
        \label{fig:loss_fhn_fno_un}
    \end{subfigure}%
    \hfill%
    \begin{subfigure}{0.48\textwidth}
        \centering
        \includegraphics[width=\textwidth]{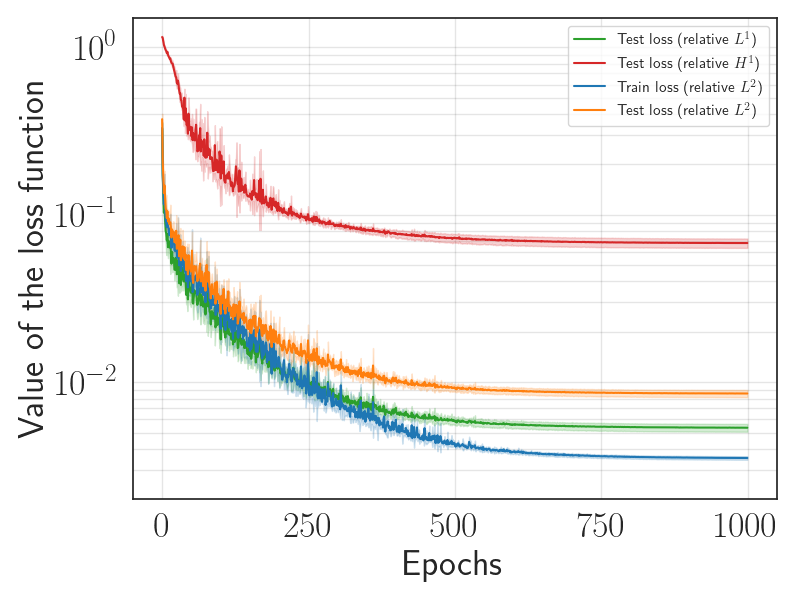}
        \caption{Loss values for the constrained FNO.}
        \label{fig:loss_fhn_fno_con}
    \end{subfigure}
    \caption{FHN model: FNO training relative $L^2$ (blue) loss and relative test $L^1$ (green), $L^2$ (orange), and $H^1$ (red) loss functions. (a) Unconstrained FNO model, (b) constrained FNO model.}
    \label{fig:loss_FHN_FNO}
\end{figure}
As shown in Fig. \ref{fig:loss_FHN_FNO}, the relative $L^1$ and $L^2$ norms yield similar accuracies, while the $H^1$ norm is about an order of magnitude higher. This discrepancy is primarily due to the fact that the training methodology employs the $L^2$ norm rather than a derivative-based norm like the $H^1$ norm. The higher values in the $H^1$ norm indicate that while the model accurately captures the function values, it exhibits larger errors when the derivative is taken into consideration. Despite achieving similar accuracy in the final results, the unconstrained network demonstrates greater computational efficiency, requiring roughly half as many training epochs as the constrained architecture to converge to comparable error levels. The constraint of a fixed number of parameters potentially can improve model efficiency by creating lightweight architectures. However, this advantage is offset by the increased number of training epochs required for convergence.
\\
To provide a more comprehensive understanding of the error distribution, we present both bar and box plots in Fig. \ref{fig:fhn_statistics}. The bar plots illustrate the mean errors across the test dataset, while the box plots reveal the variability and presence of potential outliers in the error distribution, offering insights into the robustness of both architectural approaches.
\\
In Fig. \ref{figure:fhn grafici}, we show a detailed comparison between the unconstrained Fourier Neural Operator and the solution given by the Runge-Kutta method, which serves as our high-fidelity solution. The visual comparison confirms the quantitative results and demonstrates the ability of the FNO to accurately capture the complex dynamics of the FitzHugh-Nagumo system over different applied currents.
\begin{figure}[!ht]
    \centering
    \begin{subfigure}[t]{0.49\textwidth}
        \centering
        \includegraphics[width=0.9\textwidth]{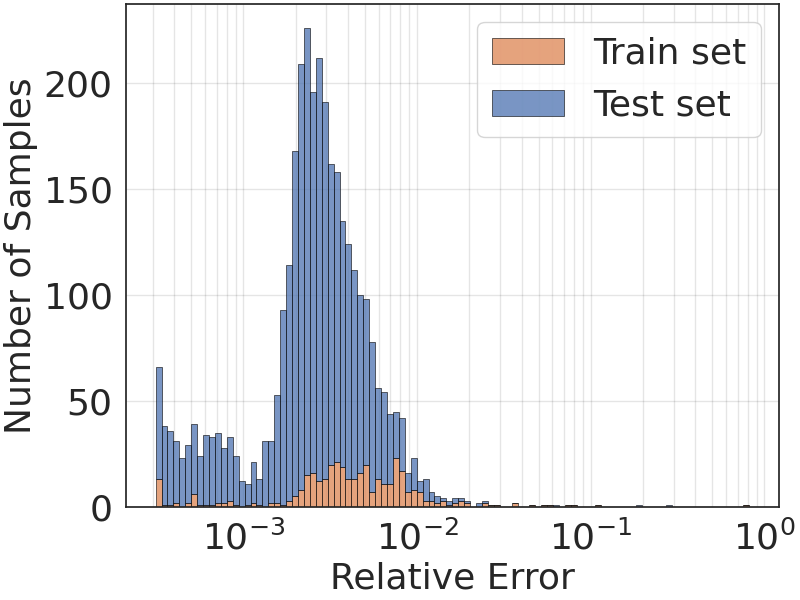}
        \caption{Unconstrained FNO bar plot.}
        \label{fig:barplot_fhn_unconstrained}
    \end{subfigure}%
    \hfill%
    \begin{subfigure}[t]{0.49\textwidth}
        \centering
        \includegraphics[width=0.9\textwidth]{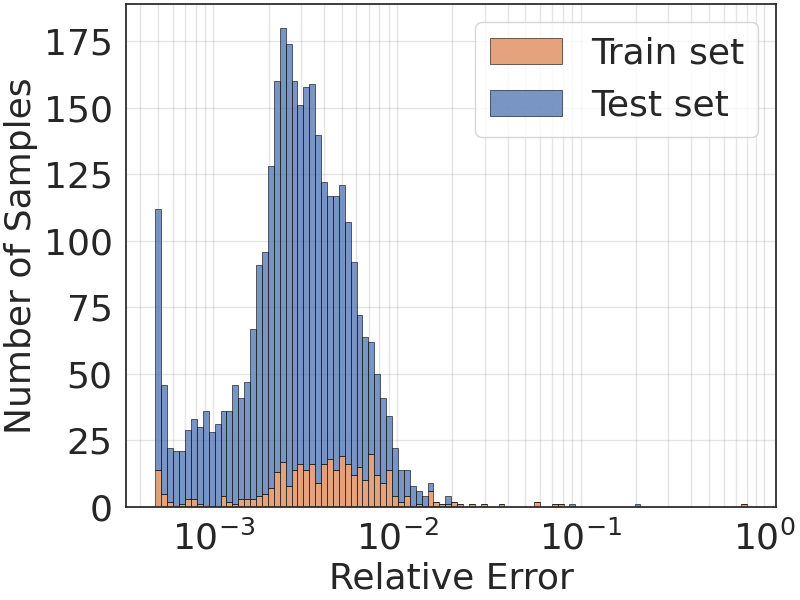}
        \caption{Constrained FNO bar plot.}
        \label{fig:barplot_fhn_constrained}
    \end{subfigure}
\end{figure}
\begin{figure}[!ht]
    \ContinuedFloat 
    \centering
    \begin{subfigure}[t]{0.55\textwidth}
        \centering
        \includegraphics[width=\textwidth]{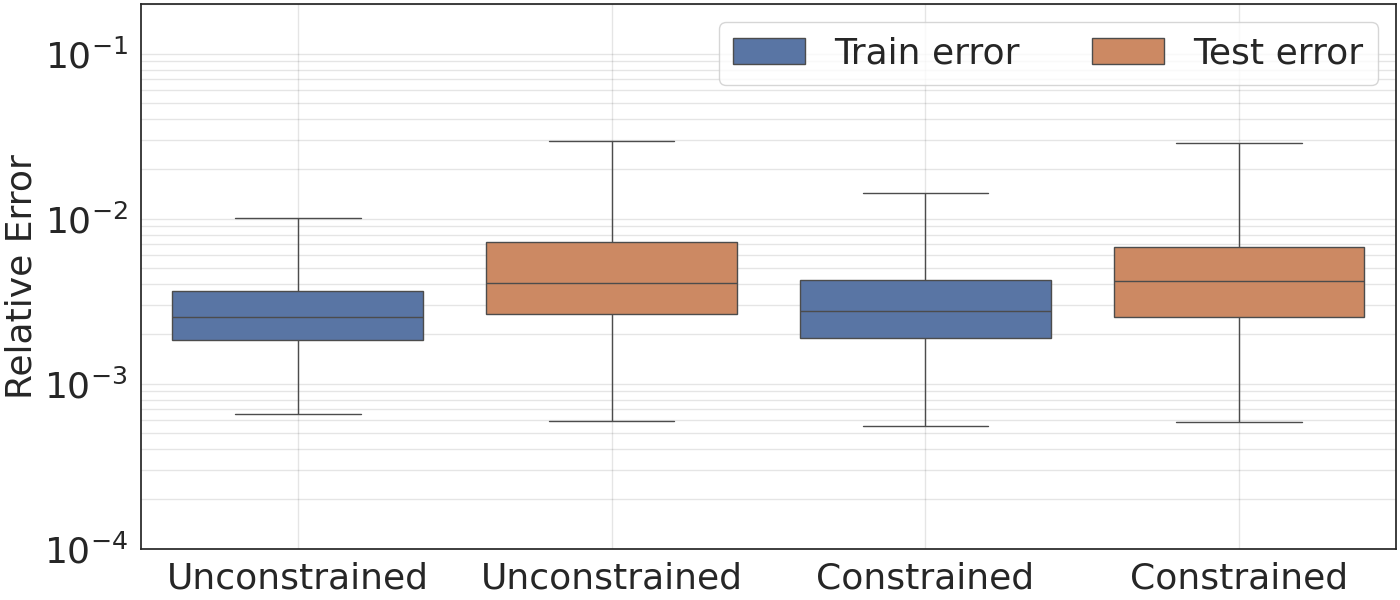}
        \caption{Box plot for unconstrained and constrained FNO.}
        \label{fig:boxplot_fhn}
    \end{subfigure}
    \caption{FHN model: FNO performance comparison. Figure \ref{fig:barplot_fhn_unconstrained} shows the bar plot of the relative $L^2$ error for the unconstrained FNO. Figure \ref{fig:barplot_fhn_constrained} shows the bar plot of the relative $L^2$ error for the constrained FNO. Figure \ref{fig:boxplot_fhn} is a box plot illustrating the distribution of relative $L^2$ errors for both the constrained and unconstrained architectures.}
    \label{fig:fhn_statistics}
\end{figure}

\begin{figure}[ht]
    \centering
    \includegraphics[width=\textwidth]{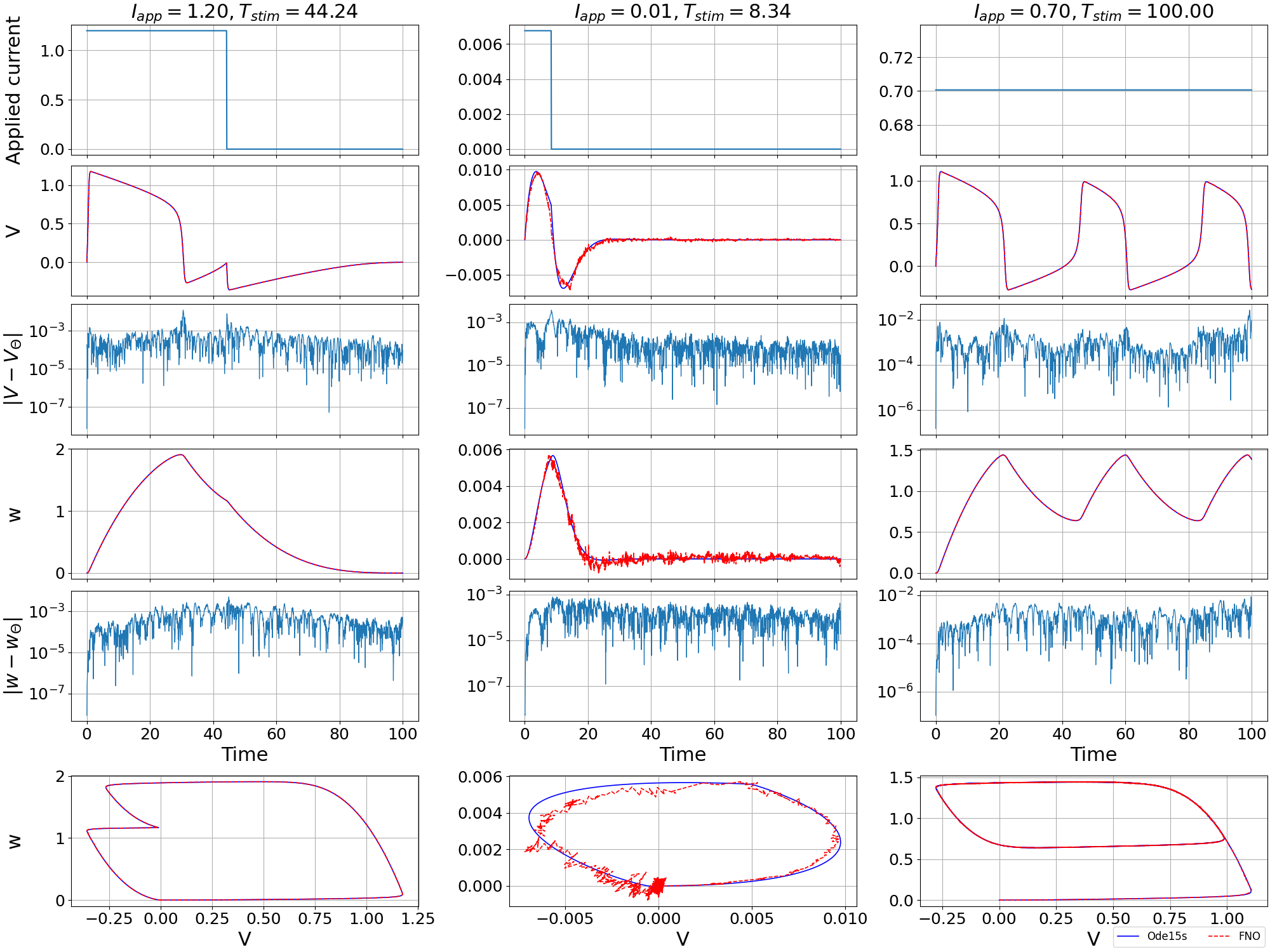}
    \addtocounter{figure}{+1}
    \caption{FHN model: examples of FNO performance. Each column represents a single example from a subset of the test dataset (Table \ref{table:fhn_dataset}). Within each column, the rows illustrate: applied current $I_{app}$, voltage $V$, pointwise error for the voltage $V$, recovery variable $w$, pointwise error for the recovery variable $w$, and phase space.}
    \label{figure:fhn grafici}
\end{figure}
\clearpage
\subsection{Hodgkin-Huxley results}
The second case considered, the Hodgkin-Huxley model, serves as a benchmark for ionic models, given its significance in the literature as highlighted in Section \ref{subsection:hh}. The parameters used in Equation \eqref{equation:hh} are listed in Table \ref{table:hh parameters}.
\begin{table}[ht!]
    \centering
    \begin{tabular}{lll}
        \hline
        Parameter      & Value             & Description                          \\
        \hline \hline
        $C_m$          & 1 $\mu F / $cm$^2$  & Membrane Capacitance                 \\
        $\bar{g}_{\Na}$ & 120    mS/cm$^2$ & Maximum $\Na^+$ channel conductance   \\
        $\bar{g}_{K}$  & 36     mS/cm$^2$ & Maximum $\K^+$ channel conductance    \\
        $\bar{g}_{L}$  & 0.3    mS/cm$^2$ & Maximum leak channel conductance     \\
        $V_{\Na}$       & 115 mV            & Nernst potential for $\Na^+$ channels \\
        $V_{K}$        & -12 mV            & Nernst potential for $\K^+$  channels \\
        $V_{L}$        & 10.6 mV           & Nernst potential for leak channels   \\
        $T$            & 100 ms            & Time                                 \\
        $V_0$          & 2.757e-02 mV      & Initial membrane potential           \\
        $m_0$          & 5.2934e-02        & Initial $\Na^+$ current activation    \\
        $h_0$          & 5.9611e-01        & Initial $\Na^+$ current inactivation  \\
        $n_0$          & 3.1768e-01        & Initial $\K^+$ current activation     \\
        \hline
    \end{tabular}
    \caption{HH model parameters, see \eqref{equation:hh}.}
    \label{table:hh parameters}
\end{table}

Similar to the previous example, the dataset consists of consists of 3000 training, 375 test, and 375 for validation examples, generated using a Runge-Kutta method tailored for stiff ODEs \cite{shampine1999solving}. Due to the increased complexity of the Hodgkin-Huxley model's dynamics compared to the FitzHugh-Nagumo model, the datasets is divided into four subsets. The ranges of intensity ($i$) and stimulus duration ($T_{stim}$) for each subset are presented in Table \ref{table:hh_dataset} for training, testing, and validation, respectively.
 \begin{table}[ht!]
    \centering
    \begin{tabular}{cccccc}
        \hline
        \multicolumn{6}{c}{\textbf{Training Dataset}} \\
        \hline
        Name       & \multicolumn{2}{c}{Range of values of $i$} & \multicolumn{2}{c}{Range of values of $T_{stim}$} & Number of examples              \\
                   & min                                        & max                                               & min                & max &      \\
        \hline\hline
        $t_0$      & -                                          & -                                                 & 0                  & 0   & 20   \\
        General    & 2                                          & 10                                                & 0.01               & 100 & 2380 \\
        nap        & 1e-4                                       & 2                                                 & 0.01               & 100 & 100  \\
        $i_{high}$ & 50                                         & 200                                               & 0.01               & 100 & 300  \\
        $t_{fin}$  & 2                                          & 30                                                & 100                & 100 & 200  \\
        \hline
        \multicolumn{6}{c}{} \\
        \hline
        \multicolumn{6}{c}{\textbf{Test and Validation Dataset}} \\
        \hline
        Name       & \multicolumn{2}{c}{Range of values of $i$} & \multicolumn{2}{c}{Range of values of $T_{stim}$} & Number of examples             \\
                   & min                                        & max                                               & min                & max &     \\
        \hline\hline
        $t_0$      & -                                          & -                                                 & 0                  & 0   & 10  \\
        General    & 2                                          & 10                                                & 0.01               & 100 & 275 \\
        nap        & 1e-4                                       & 2                                                 & 0.01               & 100 & 30  \\
        $i_{high}$ & 50                                         & 200                                               & 0.01               & 100 & 30  \\
        $t_{fin}$  & 2                                          & 30                                                & 100                & 100 & 30  \\
        \hline
    \end{tabular}
    \caption{HH model: Range of values of the training set (first table) and the test and validation dataset (second table).}
    \label{table:hh_dataset}
\end{table}

Table \ref{table:numerical_results_hh_fno} shows the results obtained using the hyperparameters from Tables \ref{table:hyper-FNO-uncostrained} and \ref{table:hyper-FNO-samedofs} for the unconstrained and constrained cases, respectively. The table includes the total number of trainable parameters (in millions), memory usage during training (GB), the training time (minutes), and the mean relative $L^2$ error (averaged over five runs with different seeds).
\begin{table}[ht!]
    \centering
    \begin{tabular}{ccccc|c}
        \hline
        Mode       & Number of & Memory  &  Time & \multicolumn{2}{c}{rel $L^2$ error}                         \\
                   &     parameters       &              &               & Training                              & Test                  \\
        \hline\hline
        constr. optim. & $0.63$ M   & $1.21$ GB    & $20$ min     & $(1.31 \pm 0.092) \%$                 & $(2.71 \pm 0.106) \%$ \\
        unconstr. optim.       & $13.44$ M  & $1.95$ GB   & $40$ min     & $(0.93 \pm 0.020) \%$                 & $(2.34 \pm 0.062) \%$ \\
        \hline
    \end{tabular}
    \caption{HH model: FNO result summary.}
    \label{table:numerical_results_hh_fno}
\end{table}

\begin{figure}[ht]
    \centering
    \begin{subfigure}{0.49\textwidth}
        \centering
        \includegraphics[width=\textwidth]{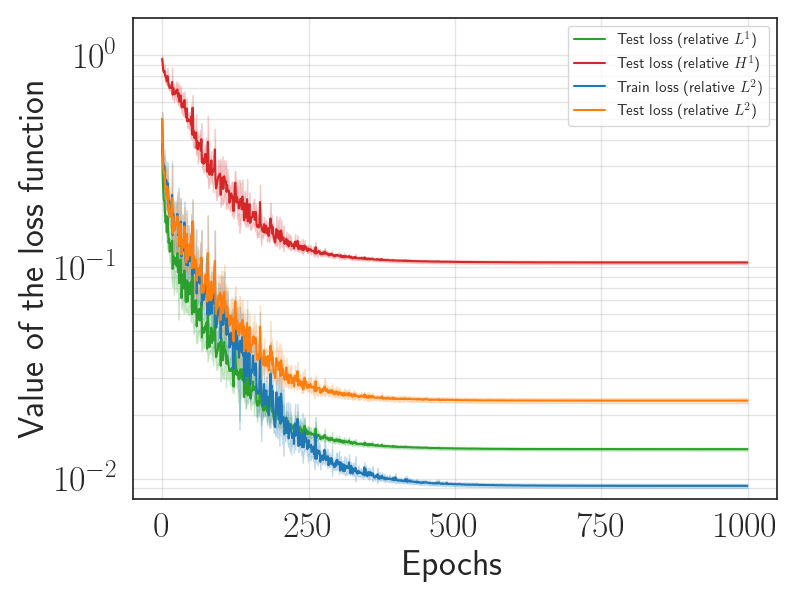}
        \caption{Loss values for the unconstrained FNO.}
    \end{subfigure}%
    \hfill%
    \begin{subfigure}{0.49\textwidth}
        \centering
        \includegraphics[width=\textwidth]{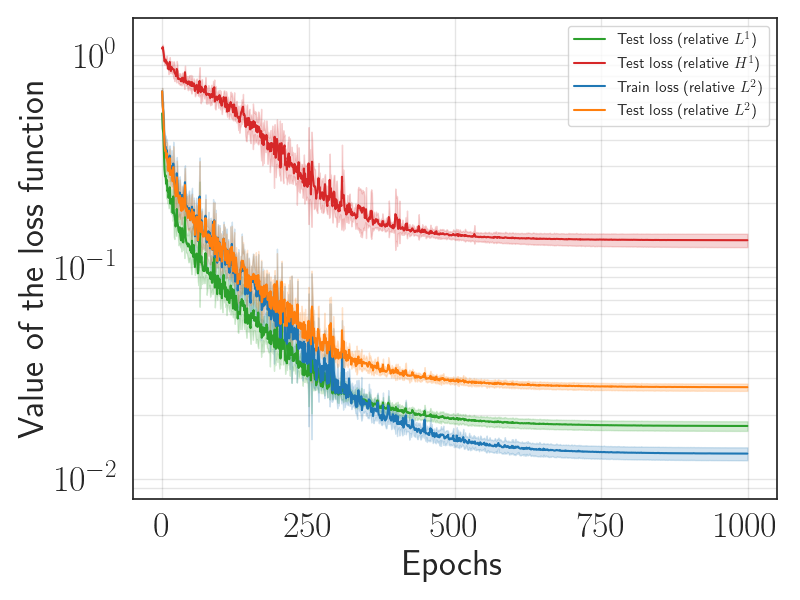}
        \caption{Loss values for the constrained FNO.}
    \end{subfigure}
    \caption{HH model: FNO training relative $L^2$ (blue) loss and test relative $L^1$ (green), $L^2$ (orange), and $H^1$ (red) loss functions. (a) Unconstrained FNO model, (b) constrained FNO model.}
    \label{fig:loss_hh_FNO}
\end{figure}

\begin{figure}
    \centering
    \begin{subfigure}[t]{0.49\textwidth}
        \centering
        \includegraphics[width=\textwidth]{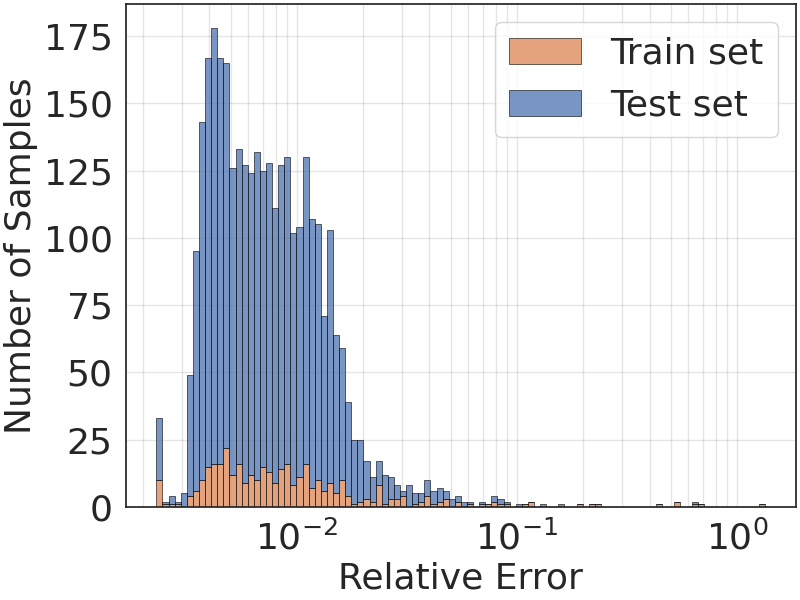}
        \caption{Unconstrained FNO bar plot.}
        \label{fig:barplot_hh_unconstrained}
    \end{subfigure}%
    \hfill%
    \begin{subfigure}[t]{0.49\textwidth}
        \centering
        \includegraphics[width=\textwidth]{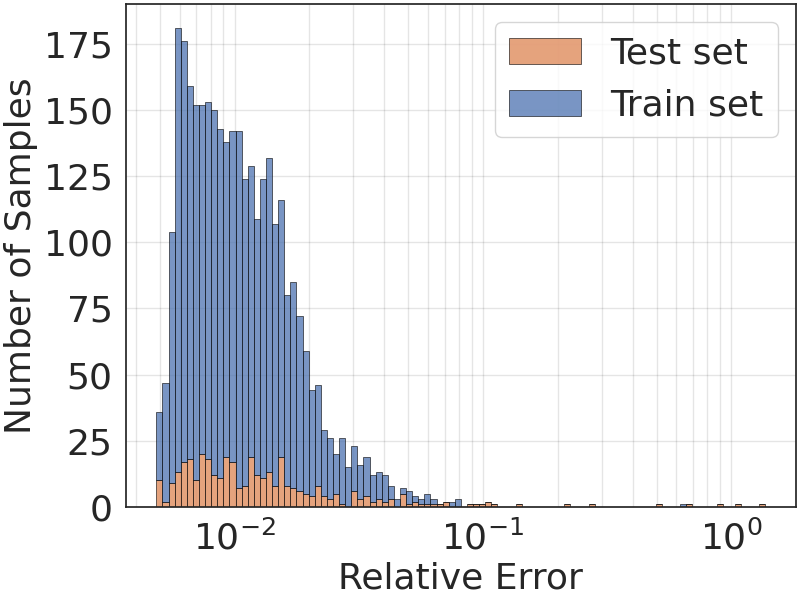}
        \caption{Constrained FNO bar plot.}
        \label{fig:barplot_hh_constrained}
    \end{subfigure}%
\end{figure}
\clearpage
\begin{figure}
\ContinuedFloat
\centering
    \begin{subfigure}[t]{0.55\textwidth}
        \centering
        \includegraphics[width=\textwidth]{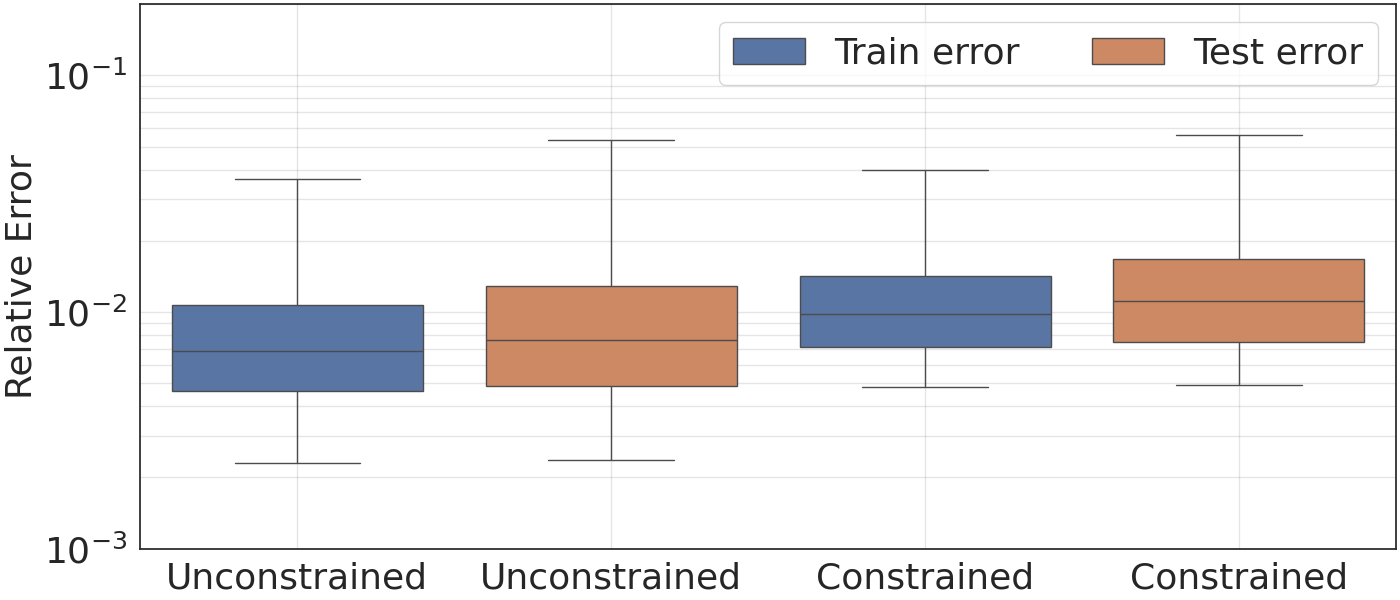}
        \caption{Box plot for the unconstrained and constrained FNO.}
        \label{fig:boxplot_hh}
    \end{subfigure}
    \caption{HH model:
         FNO performance comparison. Fig. \ref{fig:barplot_hh_unconstrained} shows the bar plot of the relative $L^2$ error for the unconstrained FNO. Fig. \ref{fig:barplot_hh_constrained} shows the bar plot of the relative $L^2$ error for the constrained FNO. Fig. \ref{fig:boxplot_hh} is a box plot illustrating the distribution of relative $L^2$ errors for both the constrained and unconstrained architectures.
    }
    \label{fig:hh_statistics}
\end{figure}

\begin{figure}[ht]
    \centering
    \includegraphics[width=\textwidth]{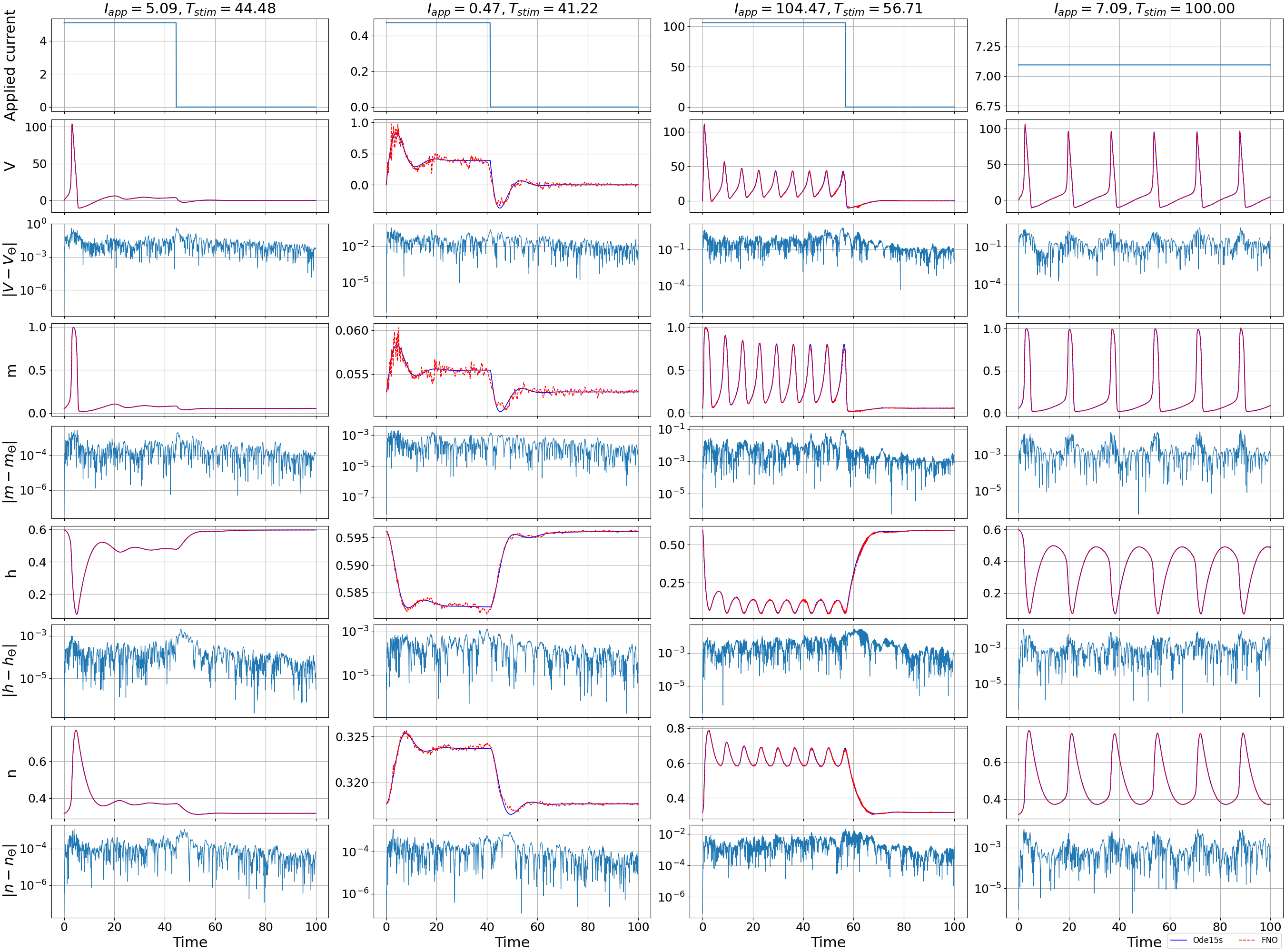}
    \caption{HH model: examples of FNO performance. Each column represents a single example from a subset of the test dataset (Table \ref{table:hh_dataset}) Within each column, the rows represent: applied current $I_{app}$, voltage $V$, point-wise error for the voltage, gating variable m, point-wise error for m, gating variable n, point-wise error for n, gating variable h, point-wise error for the gating variable h.}
    \label{figure:hh grafici}
\end{figure}

\clearpage
As for the FitzHugh-Nagumo model, from Table \ref{table:numerical_results_hh_fno} we can observe that the unconstrained architecture did not provide a significant accuracy advantage over the constrained architecture. To further evaluate and investigate the difference between the two architectures, we examined the relative $L^1$, $L^2$, and $H^1$ norms (averaged over five runs with different seeds). 
\\
From the previous Fig. \ref{fig:loss_hh_FNO} we can draw conclusions similar to those highlighted in the FitzHugh-Nagumo case. As before, we report both bar and box plots in Fig. \ref{fig:hh_statistics} to show the distribution of training and testing errors.
\\
Finally in Fig. \ref{figure:hh grafici}, we compare the prediction of the unconstrained Fourier Neural Operator with the solution given by the Runge-Kutta method.

\subsection{O'Hara-Rudy results}
The last ionic model considered is the O'Hara-Rudy model, introduced in Section \ref{subsection:Ord}. Due to the model's high dimensionality in terms of variables and parameters, a comprehensive description is not provided here but can be found in the supplementary material of \cite{o2011simulation}. The dataset, consisting of 3000 training, 375 test, and 375 for validation examples, is generated using a Runge-Kutta method tailored for stiff ODEs \cite{shampine1999solving}. To account for the model's complexity, the dataset was divided into five subsets, each representing a specific range of current intensity ($i$) and stimulus duration ($T_{stim}$).  The values are reported in Table \ref{table:ord_dataset} for training, testing, and validation, respectively.

\begin{table}[ht!]
    \centering
    \begin{tabular}{cccccc}
        \hline
        \multicolumn{6}{c}{\textbf{Training Dataset}} \\
        \hline
        Name       & \multicolumn{2}{c}{Range of values of $i$} & \multicolumn{2}{c}{Range of values of $T_{stim}$} & Number of examples               \\
                   & min                                        & max                                               & min                & max  &      \\
        \hline\hline
        $t_0$      & -                                          & -                                                 & 0                  & 0    & 50   \\
        General    & 0                                          & 20                                                & 2                  & 5    & 2050 \\
        $i_{Mid}$  & 1.1                                        & 10                                                & 2                  & 10   & 300  \\
        $i_{low}$  & 0                                          & 1.1                                               & 0                  & 500  & 300  \\
        $t_{fin}$  & 0.7                                        & 1.1                                               & 500                & 500  & 300  \\
        \hline
        \multicolumn{6}{c}{} \\
        \hline
        \multicolumn{6}{c}{\textbf{Test and Validation Dataset}} \\
        \hline
        Name       & \multicolumn{2}{c}{Range of values of $i$} & \multicolumn{2}{c}{Range of values of $T_{stim}$} & Number of examples               \\
                   & min                                        & max                                               & min                & max  &      \\
        \hline\hline
        $t_0$      & -                                          & -                                                 & 0                  & 0    & 15   \\
        General    & 0                                          & 20                                                & 2                  & 5    & 270 \\
        $i_{Mid}$  & 1.1                                        & 10                                                & 2                  & 10   & 30  \\
        $i_{low}$  & 0                                          & 1.1                                               & 0                  & 500  & 30  \\
        $t_{fin}$  & 0.7                                        & 1.1                                               & 500                & 500  & 30  \\
        \hline
    \end{tabular}
    \caption{ORd model: Range of values of the training set (first table) and the test and validation dataset (second table).}
    \label{table:ord_dataset}
\end{table}
Table \ref{table:numerical_results_ord_fno} reports the results obtained using the hyperparameters from Tables \ref{table:hyper-FNO-uncostrained} and \ref{table:hyper-FNO-samedofs} for the unconstrained and constrained cases, respectively. The table includes the total number of trainable parameters (in millions), memory usage during training (GB), the training time (minutes), and the mean relative $L^2$ error (averaged over five runs with different seeds).
\begin{table}[ht]
    \centering
    \begin{tabular}{ccccc|c}
        \hline
        Mode       & Number of  & Memory  &  Time & \multicolumn{2}{c}{rel $L^2$ error}          \\
                   &       parameters     &              &               & Training                              & Test   \\
        \hline\hline
        constr. optim. & $0.51$ M      & $1.56$ GB       & $72$ min      & $(1,31\pm 0.03) \%$                                & $(2.42 \pm 0.03) \%$ \\
        unconstr. optim.       & $12.4$ M      & $5.41$ GB       & $187$ min      & $(0.88 \pm 0.02) \%$                                & $(2.19 \pm 0.04) \%$ \\
        \hline
    \end{tabular}
        \caption{ORd model: FNO result summary.}
        \label{table:numerical_results_ord_fno}
\end{table}

\begin{figure}[ht!]
    \centering
    \begin{subfigure}{0.49\textwidth}
        \centering
        \includegraphics[width=\textwidth]{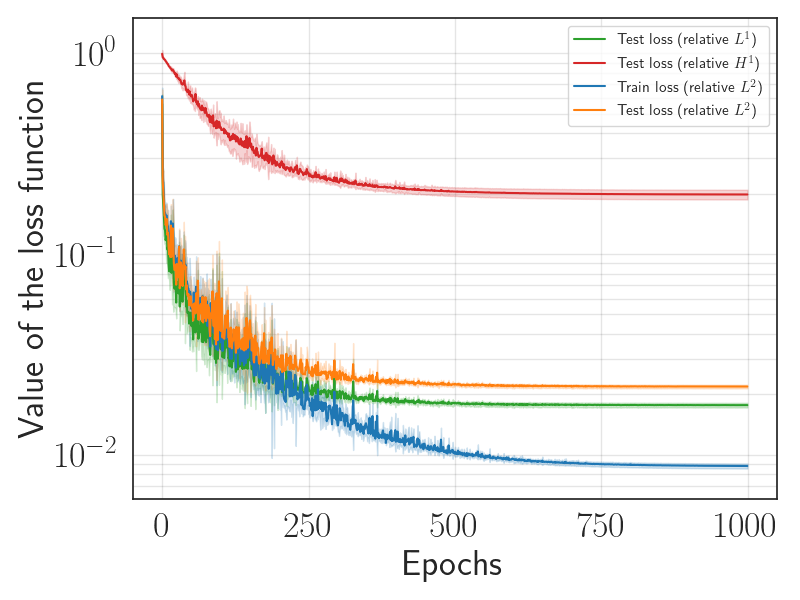}
        \caption{Loss values for the unconstrained FNO.}
    \end{subfigure}%
    \hfill%
    \begin{subfigure}{0.49\textwidth}
        \centering
        \includegraphics[width=\textwidth]{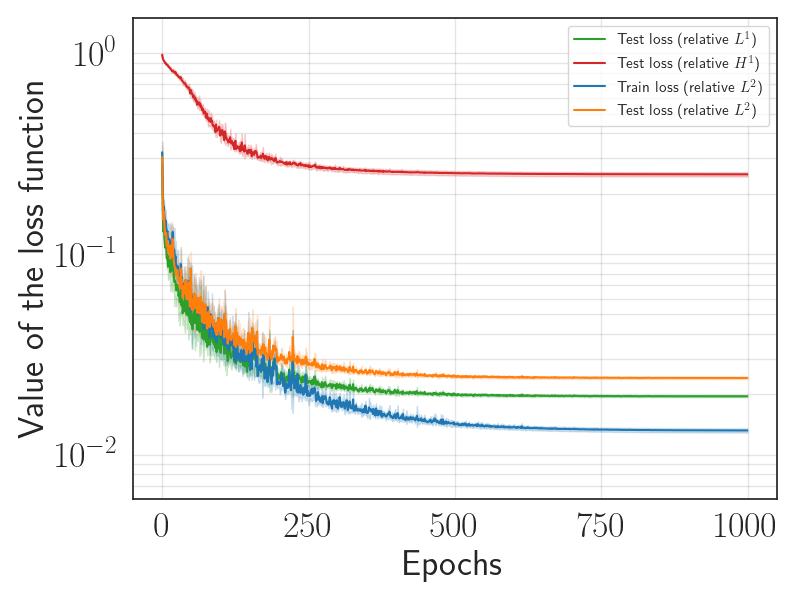}
        \caption{Loss values for the constrained FNO}
    \end{subfigure}
    \caption{ORd model: FNO training relative $L^2$ (blue) loss and test relative $L^1$ (green), $L^2$ (orange), and $H^1$ (red)
loss functions. (a) Unconstrained FNO model, (b) constrained FNO model.}
    \label{fig:loss_ORD_FNO}
\end{figure}
\newpage
\begin{figure}[!ht]
    \centering
    \begin{subfigure}[t]{0.49\textwidth}
        \centering
        \includegraphics[width=0.9\textwidth]{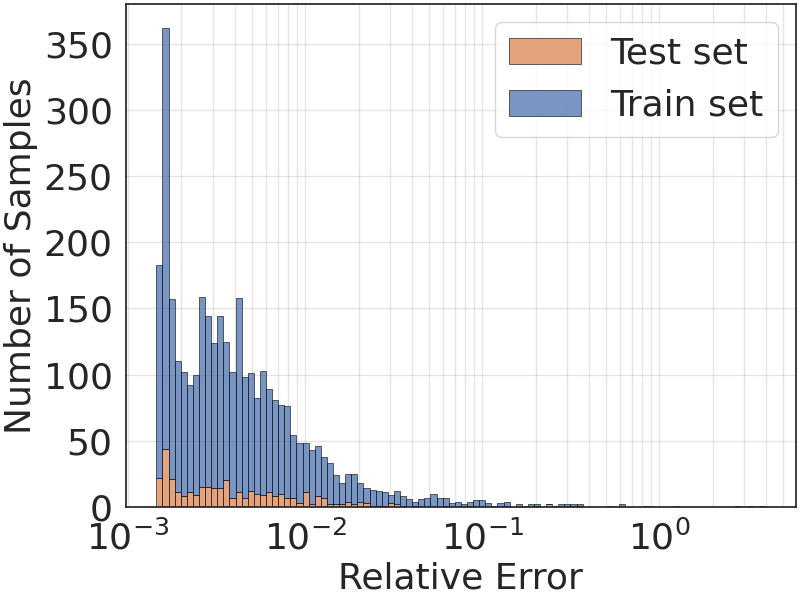}
        \caption{Unconstrained FNO bar plot.}
        \label{fig:barplot_ord_unconstrained}
    \end{subfigure}%
    \hfill%
    \begin{subfigure}[t]{0.49\textwidth}
        \centering
        \includegraphics[width=0.9\textwidth]{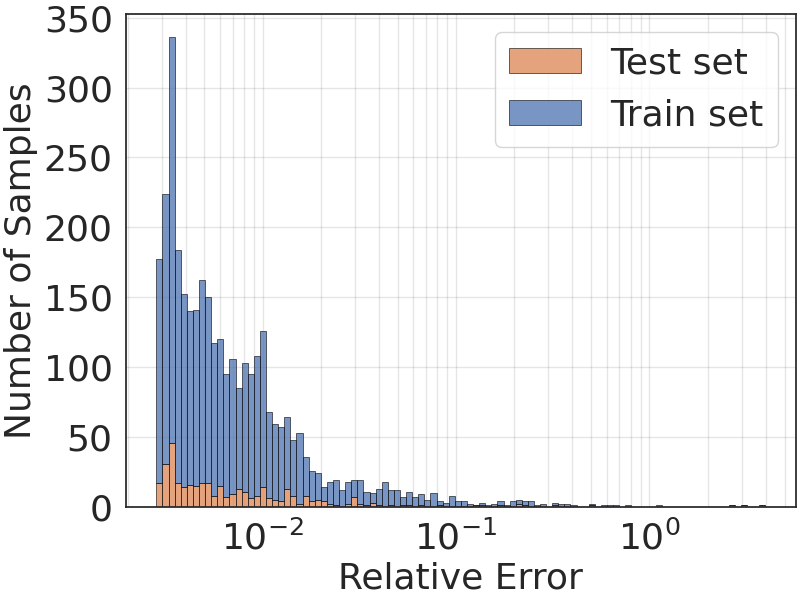}
        \caption{Constrained FNO bar plot.}
        \label{fig:barplot_ord_constrained}
    \end{subfigure}%
    \vfill%
    \vspace{.3cm}
    \begin{subfigure}[t]{0.55\textwidth}
        \centering
        \includegraphics[width=\textwidth]{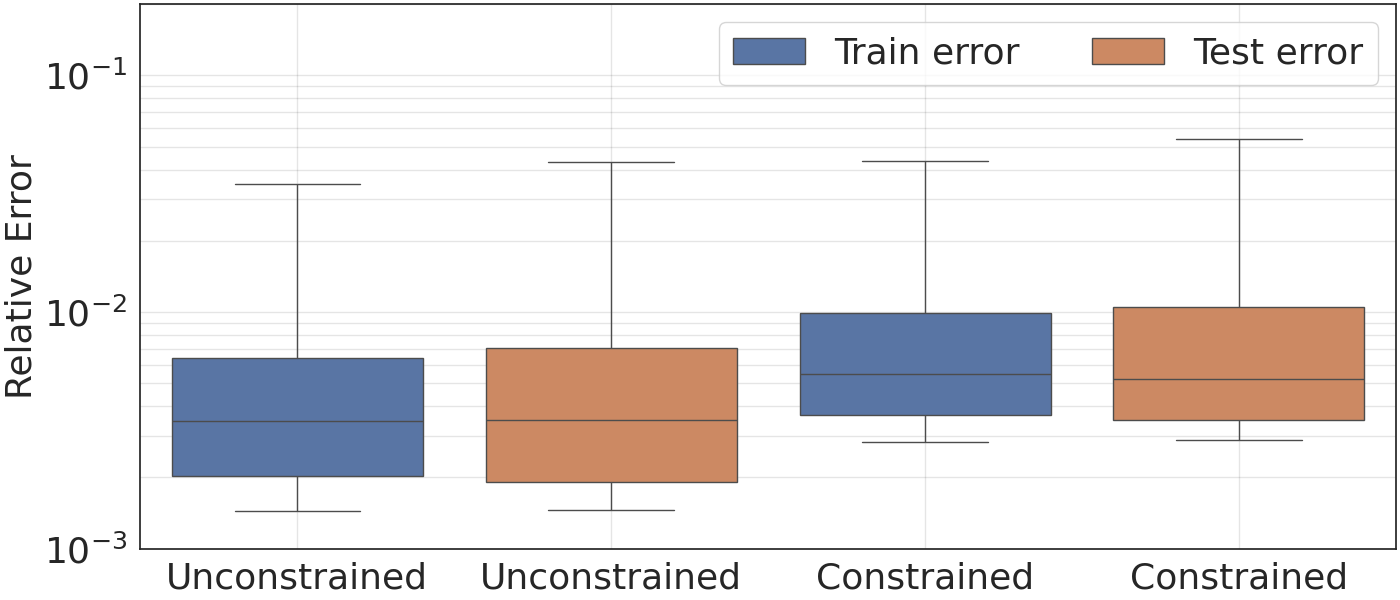}
        \caption{Box plot for the unconstrained and constrained FNO.}
        \label{fig:boxplot_ord}
    \end{subfigure}
    \hfill%
    \caption{ORd model: FNO performance comparison. Fig. \ref{fig:barplot_ord_unconstrained} shows the bar plot of the relative $L^2$ error for the unconstrained FNO. Fig. \ref{fig:barplot_ord_constrained} shows the bar plot of the relative $L^2$ error for the constrained FNO. Fig. \ref{fig:boxplot_ord} is a box plot illustrating the distribution of relative $L^2$ errors for both the constrained and unconstrained architectures.
    }
    \label{fig:ord_statistics}
\end{figure}
\begin{center}
    \begin{figure}
    \includegraphics[width=1.0\textwidth]{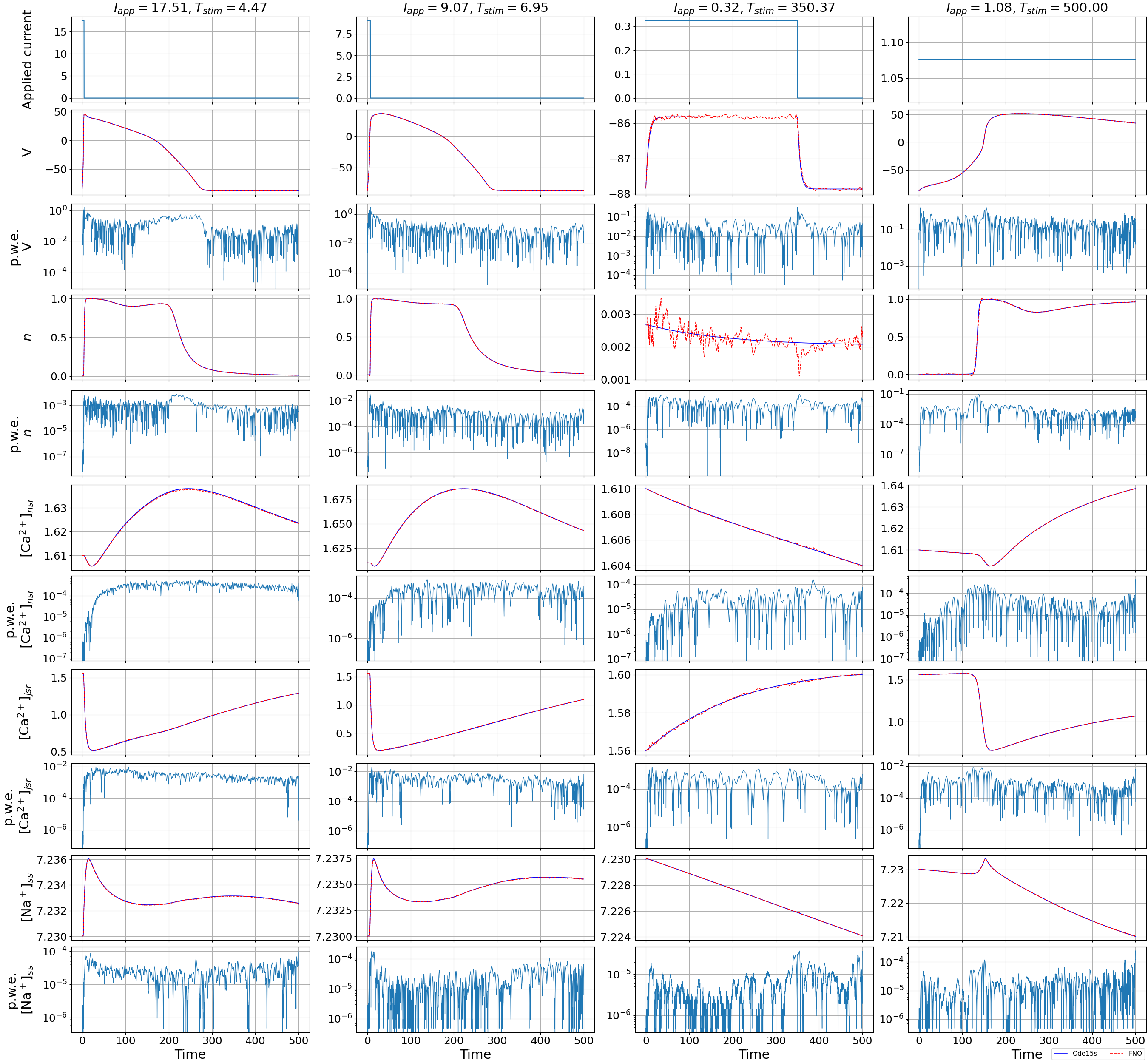}
     \caption{ORd model: examples of FNO performance. Each figure contains five plots, arranged in order from the top row as follows: current applied, voltage, point-wise error for the voltage, gating variable of $\Ca$ ($n$), point-wise error for gating variable of $\Ca$ ($n$), concentration of $[\Ca]_{nsr}$, point-wise error for the $[\Ca]_{nsr}$ concentration, concentration of $[\Ca]_{jsr}$, point-wise error for the $[\Ca]_{jsr}$ concentration, concentration of $[\Na^+]_{ss}$, point-wise error for the $[\Na^+]_{ss}$ concentration. }
    \end{figure}
\end{center}
\clearpage
As for the other models presented previously, from Table \ref{table:numerical_results_ord_fno} we can observe that the unconstrained architecture did not provide a significant accuracy advantage over the constrained architecture. In the following Fig. \ref{fig:loss_ORD_FNO}, we examined the relative $L^1$, $L^2$, and $H^1$ norms  (averaged over five runs with different seeds).
\\
Fig. \ref{fig:loss_ORD_FNO} highlights conclusions similar to those drawn for the other models. As before, the distribution of training and testing errors is shown in the bar and box plots Fig. \ref{fig:ord_statistics}.
\\
Similar to the previous examples, we compare the Runge-Kutta solutions with the unconstrained Fourier Neural Operator predictions. Due to the complexity of the O'Hara-Rudy model, we have chosen five key variables for comparison: V, n, $[\Ca]_{nsr}$, $[\Ca]_{jsr}$, and $[\Na^+]_{ss}$. These variables highlight the diverse dynamics of the model, including calcium dynamics, which is missing from previous models.

\subsection{Discussion}
\begin{figure}[!ht]
    \centering
    \begin{subfigure}[t]{0.49\textwidth}
        \centering
        \includegraphics[width=\textwidth, ]{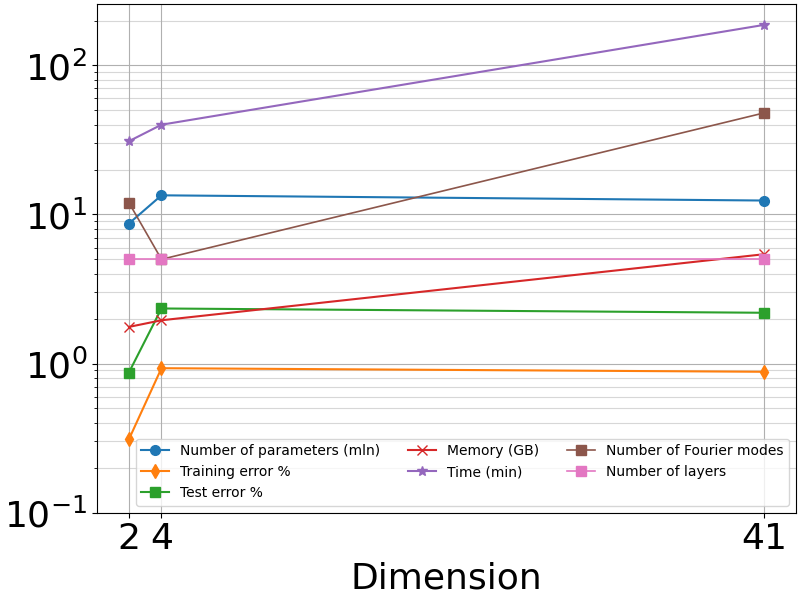}
        \caption{Unconstrained FNO}
        \label{fig:parameters_unconstrained}
    \end{subfigure}%
    \hfill%
    \begin{subfigure}[t]{0.49\textwidth}
        \centering
        \includegraphics[width=\textwidth, ]{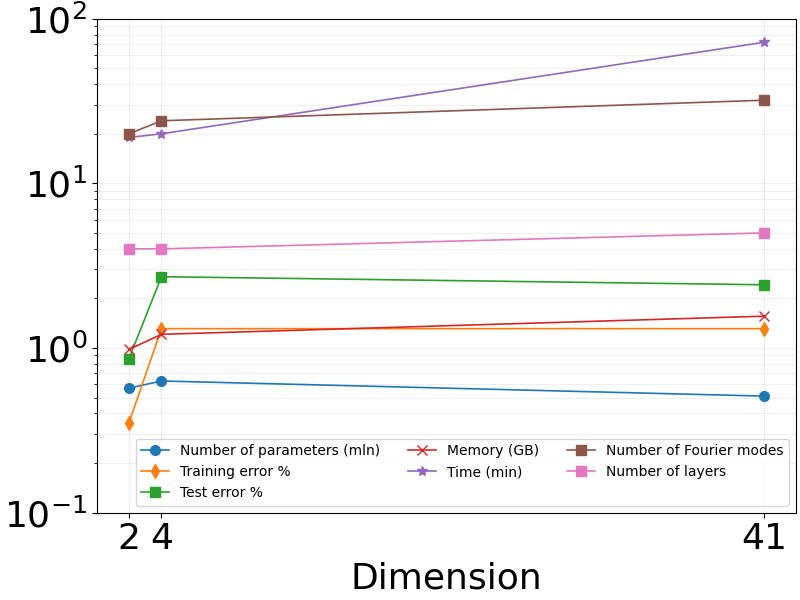}
        \caption{Constrained FNO }
        \label{fig:parameters_constrained}
    \end{subfigure}%
    \hfill%
    \caption{Dimensionality dependence of FNO parameters (from the summary Tables \ref{table:numerical_results_fhn_fno}, \ref{table:numerical_results_hh_fno}, \ref{table:numerical_results_ord_fno}).}
    
    \label{fig:parameters}
\end{figure}

\begin{figure}
    \centering
    \includegraphics[width=\textwidth]{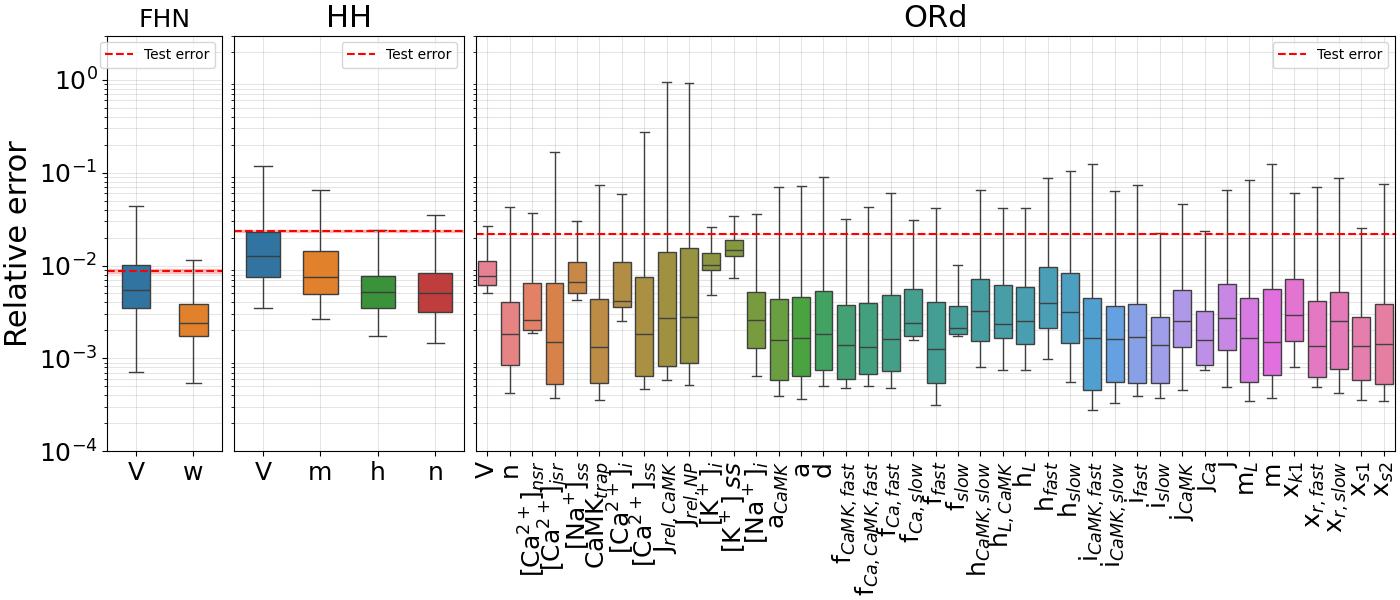}
    \caption{Component-wise box plot for the unconstrained FNO relative errors on the test set.}
    \label{fig:enter-label}
\end{figure}

In this section, we analyze how our results vary with the dimensionality of the ionic models under consideration.
We begin by investigating the dependence of several key FNO parameters on model dimensionality. These include: the number of parameters (in millions), the relative training and test errors (in $\%$), memory consumption (in GB), training time (in minutes), the number of Fourier modes, and the number of layers. In Fig. \ref{fig:parameters}, each of these quantities is reported as a function of the system dimension.
We observe that except an initial increase from the FHN (2 dimensions) to the HH (4 dimensions) model, the number of FNO parameters, relative errors, and number of layers remain bounded when the dimensions increase to 41 for the ORd model. Instead, for both constrained and unconstrained FNO architectures, the computational time required for the FNO training increases with the number of dimensions, although not exponentially. We also observe an analogous increase in memory consumption for unconstrained FNO. This increase is not directly correlated with the number of trainable parameters because, as shown in Fig. \ref{fig:parameters_unconstrained}, the number of trainable parameters does not increase with the dimension of the system. This suggests that FNOs can learn even high-dimensional systems without requiring excessively large networks in terms of trainable parameters. Moreover, the fact that the relative errors do not increase for high dimensions highlights the potential of FNOs for use in complex, high-dimensional systems. Another key aspect of analyzing multidimensional systems is understanding the distribution of error across all variables. Fig. \ref{fig:enter-label} shows the component-wise box plot of the relative test errors of unconstrained FNO for the three ionic models.
We observe that in all models, the relative test error in each variable is comparable and bounded by the global test error (red dashed line). 
The transmembrane potential $V$ seems to dominate the errors in low dimensions, but this is not the case in high dimensions, where the Potassium ($K^+$) variables contribute a higher relative error. Nevertheless, these differences are not significative, as all relative errors fall within or near the interval $[10^{-3}, \ 10^{-2}]$. This shows that FNOs accurately capture the full dynamics of complex ionic model and not just a dynamics subset of the variables.

\section{Conclusions}\label{section:conclusions}
In this study, we have successfully demonstrated the efficacy of Fourier Neural Operators in learning the complex, high-dimensional dynamics of ionic models. To this end, we evaluated FNO performance on three models of increasing complexity: the two-variable FitzHugh-Nagumo model, the four-variable Hodgkin-Huxley model, and the forty-one-variable O'Hara-Rudy model. By comparing unconstrained and constrained network architectures, we observed that while the number of parameters affects the speed of training convergence, it does not substantially change the accuracy. The FNO training time and the unconstrained FNO memory usage seem to grow at most linearly with the dimensionality of the ionic model, while the FNO parameters and relative errors seem to be bounded above independently of the dimensionality of the model. This finding suggests that FNOs can achieve robust performance even with relatively lean parameterizations, provided that sufficient training epochs are employed. The relative $L^2$ errors increased with the dimensionality of the ionic models, in particular we obtained approximately $0.8\%$ for the FHN model, $2\%$ for the HH model, and  $2\%$ for the ORd model. The successful application of FNOs to the challenging ORd model, with its numerous variables and complex dynamics, suggests their potential utility in investigating other operators arising from stiff ionic models. Such operators are of particular interest in personalized medicine, where efficient simulations of complex biological systems are essential. This work establishes FNOs as a promising tool for capturing and predicting the intricate behavior of ionic models, even in high-dimensional scenarios. Future research will explore the incorporation of spatial components into these models, transitioning from ODE systems to PDE systems. This extension will allow the simulation of spatially distributed phenomena, further expanding the applicability of FNOs in biological modeling and personalized medicine applications.

\section*{CRediT authorship contribution statement}
    \textbf{Luca Pellegrini} Conceptualization, Methodology, Software, Validation, Formal analysis, Investigation, Data Curation, Writing - Original Draft, Writing - Review \& Editing, Visualization. \textbf{Massimiliano Ghiotto} Conceptualization, Methodology, Software, Validation, Formal analysis, Investigation, Data Curation, Writing - Original Draft, Writing - Review \& Editing, Visualization. \textbf{Edoardo Centofanti} Conceptualization, Investigation, Resources, Writing - Original Draft, Writing - Review \& Editing, Visualization. \textbf{Luca Franco Pavarino} Conceptualization, Writing - Original Draft, Writing - Review \& Editing, Supervision, Project administration, Funding acquisition.

\section*{Declaration of competing interest}
The authors declare that they have no known competing financial interests or personal relationships that could have appeared to influence the work reported in this paper.

\section*{Acknowledgements}
The authors have been supported by MUR (PRIN P2022B38NR\_001) funded by European Union - Next Generation EU. EC and LFP acknowledge ISCRA for awarding our project DDO2CARD access to the LEONARDO supercomputer, owned by the EuroHPC Joint Undertaking, hosted by CINECA (Italy).

\section*{Code Availability}
The code and dataset will be made available on GitHub and Zenodo following the publication of the paper.


\begin{thebibliography}{99}

    \bibitem{Kamyar}  
        Kamyar Azizzadenesheli, Nikola , Zongyi Li, Miguel Liu-Schiaffini, Jean Kossaifi, Anima Anandkumar. \textit{Neural operators for accelerating scientific simulations and design}, Nature Reviews Physics 6, no. 5 : 320-328, 2024.
    
    
    \bibitem{bartolucci2023neural}
      Francesca Bartolucci, Emmanuel de B{\'e}zenac, Bogdan Raoni{\'c}, Roberto Molinaro, Siddhartha Mishra, Rima Alaifari,
      \textit{Are neural operators really neural operators? frame theory meets operator learning},
      SAM Research Report,
      volume 2023,
      2023.
      
    \bibitem{hyperopt11bergstra}
      James Bergstra, R{\'e}mi Bardenet, Yoshua Bengio, Bal{\'a}zs K{\'e}gl,
      \textit{Algorithms for hyper-parameter optimization},
      Advances in Neural Information Processing Systems,
      volume 24,
      2011.

    \bibitem{calvello2024continuum}
      Edoardo Calvello, Nikola B. Kovachki, Matthew E. Levine, Andrew M. Stuart,
      \textit{Continuum attention for neural operators},
      arXiv preprint arXiv:2406.06486,
      2024.

    \bibitem{cebrian2024six}
      Daniel Cebr{\'\i}an-Lacasa, Pedro Parra-Rivas, Daniel Ruiz-Reyn{\'e}s, Lendert Gelens,
      \textit{Six decades of the FitzHugh-Nagumo model: A guide through its spatio-temporal dynamics and influence across disciplines},
      arXiv preprint arXiv:2404.11403,
      2024.

    \bibitem{centofanti2024learning}
      Edoardo Centofanti, Massimiliano Ghiotto, Luca F. Pavarino,
      \textit{Learning the Hodgkin--Huxley model with operator learning techniques},
      Computer Methods in Applied Mechanics and Engineering,
      volume 432,
      page 117381,
      2024.
      
    \bibitem{cineca_leonardo}
      CINECA Supercomputing Centre, SuperComputing Applications and Innovation Department,
      \textit{LEONARDO: A Pan-European Pre-Exascale Supercomputer for HPC and AI applications},
      Journal of Large-Scale Research Facilities,
      8,
      A186,
      2024.

    \bibitem{cuomo2022scientific}
      Salvatore Cuomo, Vincenzo S. Di Cola, Fabio Giampaolo, Gianluigi Rozza, Maziar Raissi, Francesco Piccialli,
      \textit{Scientific machine learning through physics--informed neural networks: Where we are and what’s next},
      Journal of Scientific Computing,
      92,
      3,
      88,
      2022.

    \bibitem{fitzhugh1961impulses}
      Richard FitzHugh,
      \textit{Impulses and physiological states in theoretical models of nerve membrane},
      Biophysical Journal,
      1,
      6,
      445--466,
      1961.
    
   

    \bibitem{franzone2014mathematical}
      Piero C. Franzone, Luca F. Pavarino, Simone Scacchi,
      \textit{Mathematical Cardiac Electrophysiology},
      13,
      Springer,
      2014.

    \bibitem{ghiotto2025hypernosautomatedparallellibrary}
      Massimiliano Ghiotto,
      \textit{HyperNOs: Automated and Parallel Library for Neural Operators Research},
      arXiv preprint arXiv:2503.18087,
      2025.

    \bibitem{goswami2023physics}
      Somdatta Goswami, Aniruddha Bora, Yue Yu, George E. Karniadakis,
      \textit{Physics-informed deep neural operator networks},
      Machine Learning in Modeling and Simulation: Methods and Applications,
      219--254,
      2023.

    \bibitem{hodgkin1952quantitative}
      Alan L. Hodgkin, Andrew F. Huxley,
      \textit{A quantitative description of membrane current and its application to conduction and excitation in nerve},
      The Journal of Physiology,
      117,
      4,
      500,
      1952.

     \bibitem{izhikevich2007dynamical}
      Eugene M. Izhikevich,
      \textit{Dynamical Systems in Neuroscience},
        MIT press,
        2007.

    \bibitem{keener2009mathematical2}
      James Keener, James Sneyd,
      \textit{Mathematical Physiology: II: Systems physiology},
       Springer,
       2009.

    \bibitem{neuraloperator21kov}
      Nikola Kovachki, Zongyi Li, Burigede Liu, Kamyar Azizzadenesheli, Kaushik Bhattacharya, Andrew Stuart, Anima Anandkumar,
      \textit{Neural operator: Learning maps between function spaces with applications to pdes},
      Journal of Machine Learning Research,
      24,
      89,
      1--97,
      2023.


    \bibitem{lanthaler2023nonlocal}
      Samuel Lanthaler, Zongyi Li, Andrew M. Stuart,
      \textit{Nonlocality and nonlinearity implies universality in operator learning},
      arXiv preprint arXiv:2304.13221,
      2023.
    
    \bibitem{AHSHA}Liam Li, Kevin Jamieson, Afshin Rostamizadeh,      
    Ekaterina Gonina, Jonathan Ben-Tzur, Moritz Hardt, Benjamin Recht, and Ameet Talwalkar. \textit{A system for massively parallel hyperparameter tuning}, Proceedings of machine learning and systems 2: 230-246 2020.
    
    \bibitem{li2022transformer}
      Zijie Li, Kazem Meidani, Amir B. Farimani,
      \textit{Transformer for partial differential equations' operator learning},
      arXiv preprint arXiv:2205.13671,
      2022.
      
   
    \bibitem{FNO20li}
      Zongyi Li, Nikola Kovachki, Kamyar Azizzadenesheli, Burigede Liu, Kaushik Bhattacharya, Andrew M. Stuart, Anima Anandkumar,
      \textit{Fourier neural operator for parametric partial differential equations},
      arXiv preprint arXiv:2010.08895,
      2020.

    

    \bibitem{liaw2018tune}
      Richard Liaw, Eric Liang, Robert Nishihara, Philipp Moritz, Joseph E. Gonzalez, Ion Stoica,
      \textit{Tune: A Research Platform for Distributed Model Selection and Training},
      arXiv preprint arXiv:1807.05118,
      2018.
    
      
    \bibitem{loshchilov2017decoupled}
      Ilya Loshchilov, Frank Hutter,
      \textit{Decoupled weight decay regularization},
      arXiv preprint arXiv:1711.05101,
      2017.

    \bibitem{DON21lu}
      Lu Lu, Pengzhan Jin, Guofei Pang, Zhongqiang Zhang, George E. Karniadakis,
      \textit{Learning nonlinear operators via DeepONet based on the universal approximation theorem of operators},
      Nature Machine Intelligence,
      3,
      3,
      218--229,
      2021.

    \bibitem{o2011simulation}
      Thomas O'Hara, L{\'a}szl{\'o} Vir{\'a}g, Andr{\'a}s Varr{\'o}, Yoram Rudy,
      \textit{Simulation of the undiseased human cardiac ventricular action potential: model formulation and experimental validation},
      PLoS Computational Biology,
      7,
      5,
      e1002061,
      2011.

    \bibitem{raissi2019physics}
        Maziar Raissi, Paris Perdikaris, George E. Karniadakis,\textit{Physics-informed neural networks: A deep learning framework for solving forward and inverse problems involving nonlinear partial differential equations},Journal of Computational Physics,
        378,
        686-707,
        2019.

    \bibitem{CNO23raonic}
      Bogdan Raonic, Roberto Molinaro, Tobias Rohner, Siddhartha Mishra, Emmanuel de Bezenac,
      \textit{Convolutional neural operators},
      ICLR 2023 Workshop on Physics for Machine Learning,
      2023.

    \bibitem{regazzoni2024learning}
      Francesco Regazzoni, Stefano Pagani, Matteo Salvador, Luca Dede’, Alfio Quarteroni,
      \textit{Learning the intrinsic dynamics of spatio-temporal processes through Latent Dynamics Networks},
      Nature Communications,
      15,
      1,
      1834,
      2024.

    \bibitem{shampine1999solving}
      Lawrence F. Shampine, Mark W. Reichelt, Jacek A. Kierzenka,
      \textit{Solving index-1 DAEs in MATLAB and Simulink},
      SIAM Review,
      41,
      3,
      538--552,
      1999.

    \bibitem{shekarpaz2024splitting}
      Simin Shekarpaz, Fanhai Zeng, George E. Karniadakis, 
      \textit{Splitting Physics-Informed Neural Networks for Inferring the Dynamics of Integer-and Fractional-Order Neuron Models},
      Communications in Computational Physics,
      35,
      1,
      1--37,
      2024.

    \bibitem{tripura2023wavelet}
      Tapas Tripura, Souvik Chakraborty,
      \textit{Wavelet neural operator for solving parametric partial differential equations in computational mechanics problems},
      Computer Methods in Applied Mechanics and Engineering,
      404,
      115783,
      2023.
      
    \bibitem{wu2024transolver}
      Haixu Wu, Huakun Luo, Haowen Wang, Jianmin Wang, Mingsheng Long,
      \textit{Transolver: A fast transformer solver for PDEs on general geometries},
      arXiv preprint arXiv:2402.02366,
      2024.
      
    
    
    
    
    

    
\end{thebibliography}
\end{document}